%% file: main.tex
\documentclass{article} 
\usepackage{iclr2027_conference,times}

\input{math_commands.tex}

\usepackage{titletoc}

\usepackage{algorithm}
\usepackage{algpseudocode}

\input{macro}

\usepackage{booktabs}
\usepackage[most]{tcolorbox}
\usepackage{enumitem}
\usepackage{multirow}
\usepackage{array}
\usepackage{rotating}

\usepackage{amsmath}
\usepackage{amssymb}
\usepackage{amsthm}

\newtheorem{theorem}{Theorem}
\newtheorem*{theorem*}{Theorem}

\newtheorem{lemma}{Lemma}
\newtheorem{assumption}{Assumption}

\usepackage{subcaption}
\usepackage{graphicx}
\usepackage{wrapfig}

\usepackage[svgnames]{xcolor}
\usepackage{hyperref}
\usepackage{url}
\hypersetup{
    colorlinks=true,
    citecolor=Navy,
    linkcolor=Maroon,
    urlcolor=Orchid
}

\title{Planning with Transformers: \\ 
Chain of Computation and Structured Context Windows}

\author{
Ehsan Futuhi\textsuperscript{1} , Nathan R. Sturtevant\textsuperscript{1,2} \\
\textsuperscript{1}Department of Computing Science, University of Alberta \quad \textsuperscript{2}Alberta Machine Intelligence Institute (Amii) \\
\texttt{\{futuhi, nathanst\}@ualberta.ca}
}

\iclrfinalcopy 
\begin{document}

\maketitle

\input{sections/abstract}

\input{sections/introduction}


\input{sections/approach}

\input{sections/puzzles}

\input{sections/experiments}

\input{sections/discussion}

\input{sections/conclusion}

\newpage
\bibliography{iclr2027_conference}
\bibliographystyle{iclr2027_conference}

\appendix
\input{sections/appendix}

\end{document}

%% file: math_commands.tex
\usepackage{amsmath,amsfonts,bm}

\def\eqref#1{equation~\ref{#1}}
\def\Eqref#1{Equation~\ref{#1}}

\def\1{\bm{1}}

\DeclareMathAlphabet{\mathsfit}{\encodingdefault}{\sfdefault}{m}{sl}
\SetMathAlphabet{\mathsfit}{bold}{\encodingdefault}{\sfdefault}{bx}{n}



%% file: macro.tex
\newcommand{\sstart}{s_{\mathrm{start}}}
\newcommand{\sgoal}{s_{\mathrm{goal}}}
\newcommand{\plan}{\Pi}

\newcommand{\LM}{\mathcal{M}}
\newcommand{\memory}{SCW}
\newcommand{\pointer}{p}

\newcommand{\name}{Chain of Computation}
\newcommand{\abbname}{COC}

\newcommand{\ctx}{X_{\pointer}}

\newcommand{\ntarget}{n_{output}}
\newcommand{\ninput}{n_{prompt}}

%% file: sections/abstract.tex
\begin{abstract}

Large Language Models (LLMs) have had a remarkable impact across many areas of machine learning. However, recent studies have shown that they struggle to reliably solve planning problems, particularly when compared with classical planners and planning frameworks such as PDDL. Even with recent advances in reasoning-oriented models, the extent to which LLMs can genuinely perform planning remains an open question. At the same time, theoretical results have shown that transformers, the core architecture underlying modern LLMs, are Turing-complete. In this work, we investigate this apparent gap between the theoretical computational power of LLMs and their empirical planning performance. We propose \textbf{\name{} (\abbname{})}, a computational architecture that places a transformer-based LM inside an iterative loop and leverages its strength as a pattern-matching system rather than requiring it to generate an entire plan in a single pass. \abbname{} uses a \textbf{Structured Context Window (SCW)} as external memory, allowing the model to access only the portion of the accumulated context required at each planning step and thereby maintain an approximately constant input size. Within this architecture, the LM learns the planning policy and world model and performs the arithmetic operations required during planning. We show that relatively small LMs trained from scratch can learn planning procedures from a small number of training instances and generalize to unseen problems, achieving success rates above 99\% on BlocksWorld and the Pancake puzzle. We further show that \abbname{} can be formulated as a pushdown automaton (PDA), eliminating the need for explicit pointer operations. This formulation can solve Tower of Hanoi (TOH) instances with up to 20 disks and enables the planner to execute the depth-first search (DFS) algorithm in perfect maze environments. Theoretically, we characterize how the accumulation of irrelevant context increases planning failure probability under the standard full-context setup and how the SCW mitigates this effect.



\end{abstract}

%% file: sections/introduction.tex
\section{Introduction}

Large Language Models (LLMs) \citep{brown2020language,touvron2023llama} have demonstrated remarkable success across a wide range of applications, including conversational agents \citep{team2023gemini,achiam2023gpt}, code generation \citep{chen2021evaluating,yang2024swe}, and financial applications \citep{wu2023bloomberggpt,xiao2025trading}. In recent years, substantial effort has been devoted to improving the reasoning capabilities of LLMs through techniques such as Chain-of-Thought (COT) prompting and reinforcement learning \citep{wei2022chain,nye2021show,deepseek2025deepseekr1,li2025scout}. These advances have led to impressive performance on a variety of benchmarks, including mathematical reasoning \citep{cobbe2021training}, logical reasoning \citep{suzgun2023challenging}, and planning \citep{kokel2026acpbench}. Despite this progress, the performance of LLMs on planning tasks involving multi-step decision making remains limited. Numerous studies have shown that LLMs fail on planning problems that are relatively easy for humans to solve \citep{valmeekam2023planning,dziri2023faith,pallagani2023understanding,sel2025llms,shojaee2025illusion,kambhampati2024position}. Moreover, even when they are able to solve planning domains such as BlocksWorld, they often fail to generalize to more complex instances of the same problem, suggesting that they have not learned the underlying planning algorithm. Some have suggested that LLMs may be incapable of planning without separate logical oracles \citep{belle2026future}. This raises a fundamental question: if LLMs are capable of sophisticated reasoning in some contexts, what prevents them from achieving the same level of success in other contexts, in particular on planning tasks?

\citet{shojaee2025illusion} investigate the performance gap between state-of-the-art large reasoning models (LRMs) and their non-reasoning counterparts in multi-step decision-making tasks. Their approach incorporates a few in-context examples and explicitly provides the planning algorithm for each domain within the prompt. One of their key findings is that, beyond a certain problem complexity threshold, the performance of both reasoning and non-reasoning models collapses to nearly zero. Interestingly, increasing problem complexity does not induce additional reasoning behavior in these models; instead, performance deteriorates at a similar or even earlier point compared to simpler instances. On the other hand, a complementary line of theoretical work investigates whether LLMs are Turing-complete, i.e., whether they can realize arbitrary computational behavior equivalent to a Turing machine \citep{lewandowski2026universal,bhattamishra2020computational,perez2021attention,giannou2023looped,malach2024autoregressive,zhong2022training}. \citet{lewandowski2026universal} show, via a proof-by-simulation, that a language model can exactly simulate a universal Lag system \citep{wang1963tag}, which is known to be computationally universal. This result implies that language models are Turing-complete and, in principle, capable of general-purpose computation. Furthermore, they demonstrate that similar computational capabilities can emerge even in randomly initialized language models, suggesting that universal computation is an inherent property of the architecture, while training primarily serves to make this capability more practical to exploit. In another influential theoretical work, \citet{giannou2023looped} show that transformers, the core architecture underlying modern language models, can learn iterative algorithms such as square root and matrix inversion when executed in a loop with encoded weights. 
Despite these encouraging theoretical results, it remains unclear how such algorithmic capabilities generalize to other tasks and, more importantly, how the corresponding weights can be learned efficiently from training data.

\begin{figure}[t]
    \centering
    \includegraphics[width=\linewidth]{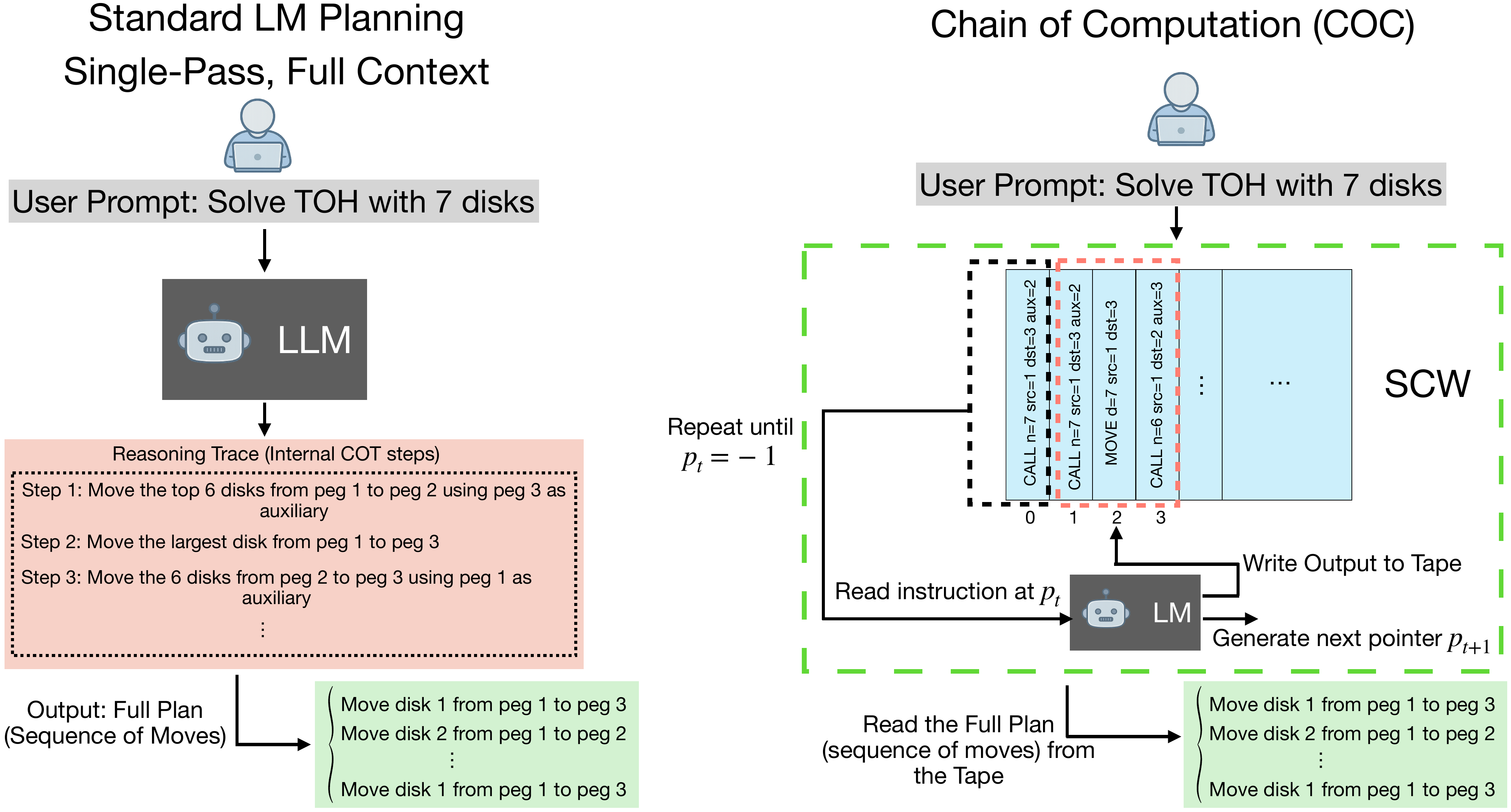}
    \caption{Comparison between standard full-context planning and Chain of Computation (COC). Instead of generating an entire plan in a single pass from the full context, COC employs a smaller LM that iteratively processes one instruction from a structured context window (SCW), updates the SCW, and generates the next pointer, $p_{t+1}$. The process terminates when $p_t=-1$, after which the complete action sequence is extracted. 
    }
    \label{fig: main-introduction}
\end{figure}

The aim of this work is to close the gap between the theoretical computational power of transformer-based language models and their empirical performance on planning tasks. Throughout this paper, we use the term \emph{LM} to refer to transformer-based language models, regardless of their size or reasoning capability. We propose \textbf{\name{} (\abbname{})}, an architecture that leverages the pattern-matching capabilities of transformers by placing them inside an iterative planning loop. The planning procedure is represented as a sequence of iterative steps, where each step consists of reading from and generating outputs to a {\bf Structured Context Window (SCW)}, which serves as an external memory. At each iteration, the LM receives the contents of a single SCW block and generates the corresponding output, which is then written to a new block in the SCW. When used for planning, the content of each block serves as an instruction for a subsequent step of the planning process. At each iteration, a \emph{pointer} selects the SCW block containing the instruction to be executed. Importantly, the pointer to the next block is generated by the LM itself, allowing the model to determine which part of the SCW to access at the next computation step. The planning process begins with the instruction stored at position zero of the SCW and terminates when the model generates the pointer value $-1$. During training, the LM is optimized on instruction--output pairs with the objective of generalizing at test time to the same operations applied to previously unseen variables and problem instances.

This paper provides extensive evaluations on three planning domains: BlocksWorld, Tower of Hanoi (TOH), and the Pancake Puzzle. In BlocksWorld and the Pancake Puzzle, we show that training on a relatively small number of problem instances at each complexity level is sufficient for the model to generalize to unseen instances across all complexity levels, achieving average success rates exceeding $99\%$. In TOH, we demonstrate that the model can generalize to unseen instructions held out during training, achieving an average success rate of $92\%$ on problems ranging from 1 to 15 disks. Analysis of the failure cases in TOH reveals that arithmetic operations involving previously unseen numbers and tokens are the primary sources of error. We show that handling these arithmetic operations separately eliminates the failures and substantially reduces the training data needed to learn to plan. Alternatively, the SCW can be implemented as a stack, where access is restricted to its top through \emph{push} and \emph{pop} operations. This allows \abbname{} to be formulated with a \emph{deterministic pushdown automaton (PDA)}, achieving the same performance without explicit pointer operations. This formulation enables the planner to execute a depth-first search (DFS) in maze environments of varying sizes. 
Our results validate empirically that LMs can be trained to plan. Finally, by decomposing the context into \emph{relevant} and \emph{irrelevant} components, we show theoretically that SCW keeps the prediction error probability fixed, while accumulated irrelevant context causes it to increase across planning iterations. We further characterize how this increase affects planning failure over long horizons.


%% file: sections/approach.tex
\section{The \name{} (\abbname{}) Architecture}

A broader review and taxonomy of prior work on LM-based planning is provided in Section~\ref{sec: realted work}, which has been moved to the appendix due to space limitations. Here, we highlight a few ideas from closely related works that motivate our approach. Scratchpads \citep{nye2021show} and chain-of-thought (COT) prompting \citep{wei2022chain} share a common underlying principle: simplifying multi-step reasoning by decomposing a problem into a sequence of next-token predictions across multiple steps. For example, if the goal is to move from room A to room B and the connecting door is locked, the first step is to pick up the key. Instead of producing the entire solution in a single pass, the reasoning process is decomposed into meaningful intermediate steps that are explicitly written to the context window. This suggests the following intuition: LMs may struggle to reliably update their internal state, but they can effectively update an external state, the context window, over long sequences. However, although updating the context is relatively easy, \citet{liu2024lost} show—using the \emph{Needle in a Haystack} test—that as the context window grows, it becomes increasingly difficult for the LM to retrieve relevant information embedded within it. This further suggests that even if the model explicitly writes all variables required for decision-making into the context at each step, it may still fail to reliably retain and utilize them as the complexity of the planning problem increases. This limitation is also observed in Searchformer \citep{lehnert2024beyond}, which trains an LM on planning traces generated by the A* algorithm \citep{hart1968formal}. The LM learns to imitate the execution of A* for each problem instance. The entire accumulated context is provided to the LM at every step, causing the context window to grow throughout the planning process. Consequently, the performance degrades as task complexity increases
(see Figure~2 in \citet{lehnert2024beyond}).


We propose the \textbf{\name{} (\abbname{})} architecture, which uses an LM as its core component for solving problems that require multi-step reasoning. The planning procedure is formulated as an iterative loop, where each iteration corresponds to a single prompt to the LM. The LM's context window is loaded from an external memory called the \textbf{Structured Context Window (SCW)}. The SCW consists of blocks, each containing the context required to execute a single instruction. At each iteration, the LM receives only the context stored in the current block and produces three outputs: (i) new instructions to be written to new blocks of the SCW, (ii) the actions at the current step, and (iii) a \emph{pointer} indicating the next instruction to be processed. To mitigate performance degradation caused by long context windows, the LM receives only the SCW block specified by the pointer, which contains all the information required to execute the corresponding instruction. To perform planning successfully, the LM must learn the planning policy, $\pi(a \mid s)$, the world model, $\mathcal{T}(s' \mid s,a)$, and the arithmetic operations required to update the pointers. An overview of the \abbname{} architecture is provided in Algorithm~\ref{alg:computational_architecture}, presented in Section~\ref{sec: coc pseudocode}. During training, each instruction is presented together with its corresponding target independently of the other instructions in the plan. The goal is to provide a finite set of training instructions that covers all three modules, enabling the LM to generalize each module to previously unseen inputs.

The pointer mechanism not only provides concise inputs to the LM but also significantly reduces computational cost. Consider a planning task that requires generating $\ntarget$ tokens. In full-context setup (e.g. \cite{lehnert2024beyond}), the model processes an input whose size grows from $\ninput$ to $\ninput+\ntarget$ as more tokens are generated, eventually approaching the maximum context length. In contrast, the proposed pointer mechanism ensures that the model receives approximately constant size inputs, determined only by the current instruction block being processed. As a result, even for long planning horizons, the computational cost per iteration remains nearly constant. A detailed empirical comparison illustrating this effect 
is presented in Section~\ref{sec: context window growth}. The pointer mechanism also reduces the probability of planning failure compared with the full-context setup. The theorem below provides lower bounds on the failure probability in both settings. The result builds on the analysis of \citet{dziri2023faith}, with the detailed proof provided in Section~\ref{sec: theoretical proofs}.

\begin{theorem}
\label{thm: main-failure-bound}
Let $f_d=g^d$ and $\widehat f_d=\widehat g^{\,d}$, where $g^d$ and $\widehat g^{\,d}$ denote the results of iteratively applying $g$ and $\widehat g$, respectively, $d$ times, and $\widehat g$ is an estimator of $g$. Let $\epsilon_0=\epsilon=\mathbb{P}(g\neq\widehat g)>0$ denote the one-step error probability when only the relevant context is provided, and let $\epsilon_t$ denote the one-step error probability at iteration $t$. Assume that the probability of recovering from an incorrect prediction is at most $c$ and that $c+\epsilon_t<1$ for all $t\geq 0$. Let $V$ denote the vocabulary size. Under CoC, where only the relevant context is retained, $\epsilon_t=\epsilon$ for all $t$, and we have $\liminf_{d\to\infty}\mathbb{P}\left(f_d\neq\widehat f_d\right)\geq 1-\frac{c}{c+\epsilon}$. Under the standard full-context setup, where irrelevant tokens accumulate across iterations, $\lim_{t\to\infty}\epsilon_t=1-\frac{1}{V}$, with $1-\frac{1}{V}-\epsilon_t=\Theta(t^{-1/2})$, and we have $\liminf_{d\to\infty}\mathbb{P}\left(f_d\neq\widehat f_d\right)\geq 1-\frac{c}{c+1-\frac{1}{V}}$.
\end{theorem}


Our approach is conceptually related to the looped transformers framework \citep{giannou2023looped}, as both employ an iterative loop to emulate the autoregressive decoding process of LMs. However, there are two key differences. First, in the looped transformers framework, the LM weights are hard-coded rather than learned. Second, that work does not provide empirical validation of the performance of such hard-coded weights.
Our approach is also different from recent methods that compress or summarize large context windows in long-horizon tasks \citep{li2025admtree,chen2025dast,liu2026autoencoding}. In these approaches, the summarization is performed either by another LM or through learned parameters called \emph{soft tokens}, both of which may introduce information loss or approximation errors. In contrast, \abbname{} uses a well-defined pointer mechanism that requires no additional learning and introduces no approximation error. Finally, the use of external memory with a neural controller was previously introduced by \citet{graves2014neural} in \emph{Neural Turing Machines}. Their architecture uses differentiable memory access and is trained end-to-end, whereas memory access in \abbname{} is symbolic.

%% file: sections/puzzles.tex
\section{Planning Domains}

\begin{figure}[t]
    \centering
    \includegraphics[width=\linewidth]{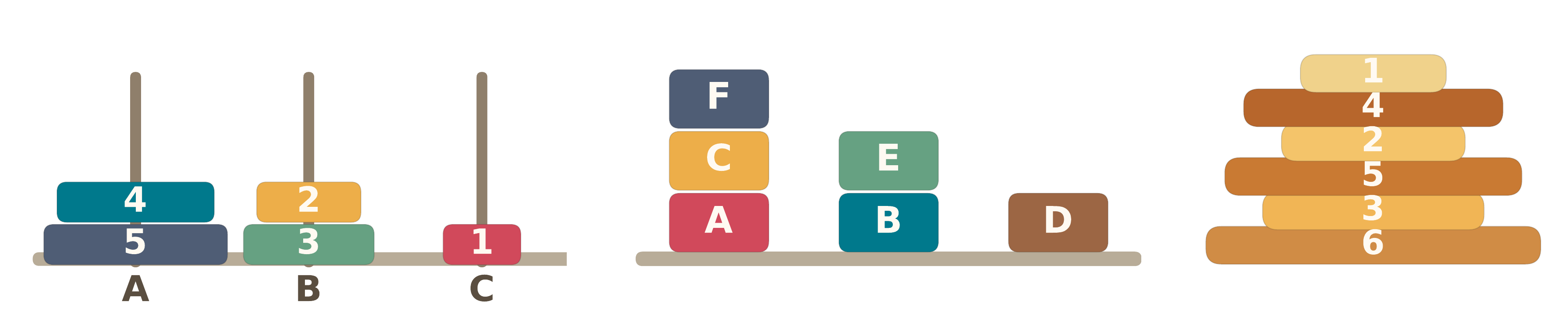}
    \caption{Planning domains used in our experiments. Left: Tower of Hanoi with $n=5$ disks. Middle: Blocksworld with $n=6$ blocks. Right: Pancake Puzzle with $n=6$ pancakes.}
    \label{fig: exp-envs}
\end{figure}

In this section, we present the planning domains used to evaluate \abbname{}. We consider two domains that are commonly used in recent work \citep{shojaee2025illusion,pallagani2023understanding,katz2024thought}: Tower of Hanoi (TOH) and BlocksWorld. In addition, we include the Pancake Puzzle \citep{GATES197947} as a classical combinatorial planning domain. These domains are illustrated in Figure~\ref{fig: exp-envs} and described below. For each domain, we show how the planning algorithm can be formulated as prompt--target pairs with pointers, enabling the training of the LM.

\paragraph{Tower of Hanoi (TOH).}
The Tower of Hanoi is a classical planning problem that requires long-horizon planning and strict adherence to constraints. The environment consists of three pegs and $n$ disks of distinct sizes, initially stacked on a single peg in decreasing order of size (largest at the bottom, smallest at the top). The objective is to move the entire stack to a target peg while satisfying two constraints: (i) only one disk can be moved at a time, and (ii) a larger disk can never be placed on top of a smaller disk. Following prior work, we define the problem complexity in terms of the number of disks $n$, as the optimal solution length grows exponentially with $n$ (i.e., $2^n - 1$ moves). 

The standard recursive planning algorithm for TOH is based on the following decomposition. To move $n$ disks from source peg $s$ to destination peg $d$ using auxiliary peg $a$, one first moves the top $n-1$ disks from $s$ to $a$, then moves disk $n$ from $s$ to $d$, and finally moves the $n-1$ disks from $a$ to $d$. Formally, if $\textsc{Hanoi}(n,s,d,a)$ denotes the solution procedure, then
\[
\textsc{Hanoi}(n,s,d,a)
=
\textsc{Hanoi}(n-1,s,a,d)
\;\circ\;
\texttt{Move}(n,s,d)
\;\circ\;
\textsc{Hanoi}(n-1,a,d,s),
\]
with the base case $\textsc{Hanoi}(1,s,d,a)=\texttt{Move}(1,s,d)$. Thus, the recursive algorithm alternates between two types of operations: recursive subproblem expansion and primitive disk moves. We represent this procedure as a stack of instructions. Each prompt to the LM corresponds to either a call or a move instruction. 
A complete example illustrating how the context window evolves with all prompts and LM outputs is provided in Appendix~\ref{sec: toh-prompt-example}. The TOH domain is unique in that the state does not need to be explicitly represented. The recursive algorithm can just generate the list of actions used to solve the problem.

\paragraph{Blocksworld.}

Blocksworld is a classical planning domain that requires reasoning over long sequences of actions with constraints on how blocks can be moved. The environment consists of a set of $n$ distinct blocks arranged in stacks on a tabletop. The goal is to transform an initial block configuration into a given target configuration. At each step, only a clear block—i.e., a block with no other block on top of it—can be moved, and it can be placed either on the table or on top of another clear block. The complexity is defined by the number of blocks $n$ and grows linearly with $n$. The planning strategy is to first unstack the blocks onto the table, and then stack them into the goal configuration. We formulate this strategy as $\textsc{BlocksWorld}(n,s,g)$, where $n$ is the number of blocks, $s$ is the current state (equal to the start configuration initially), and $g$ is the goal configuration. A complete example illustrating how the context window evolves is provided in Appendix~\ref{sec: bw-prompt-example}.



\paragraph{Pancake Puzzle.}
The Pancake Puzzle is a classical combinatorial planning problem that requires sequential decision-making under constrained actions that change the arrangement of pancakes. The environment consists of a stack of $n$ pancakes of distinct sizes in some initial order. The goal is to transform the stack into the sorted order, with the largest pancake at the bottom and the smallest at the top. The only allowed action is a prefix flip: at each step, the top $k$ pancakes, for some $1 < k \leq n$, can be flipped, reversing their order. We define the problem complexity in terms of the number of pancakes $n$, as the number of possible configurations grows factorially with $n$. The strategy for solving the Pancake Puzzle is to repeatedly apply the following process until no unsorted pancake remains: first identify the largest unsorted pancake and flip the prefix ending at its current position to bring it to the top; then flip the prefix ending at its target position to place it correctly. 




We formulate the planning problem as $\textsc{Pancake}(n,s,g)$, where $n$ is the number of pancakes, $s$ is the start configuration, and $g$ is the goal configuration, which is fixed across all problem instances. The planning procedure begins with the initial prompt $\texttt{pointer}=0\ \texttt{CALL}(s,g)$. At each step, given
$\texttt{pointer}=i\ \texttt{CALL}(s,g)$,
the LM generates, in order, $P_x$, $\texttt{FLIP}(\operatorname{cur}(P_x))$, $s'$, $\texttt{FLIP}(\operatorname{goal}(P_x))$, $s''$, and the next instruction $\texttt{pointer}=i+1\ \texttt{CALL}(s'',g)$. Here, $P_x$ denotes the largest unsorted pancake, $\operatorname{cur}(P_x)$ its current position in $s$, and $\operatorname{goal}(P_x)$ its target position in $g$. The states $s'$ and $s''$ denote the configurations obtained after the first and second flips, respectively. A complete example, including all prompts and LM outputs, is provided in Appendix~\ref{sec: pancake-prompt-example}. In Pancake Puzzle, learning the world model requires learning how to reverse a set of pancakes from the state representation, and the policy requires finding the maximum unsorted pancake.




%% file: sections/experiments.tex
\section{Experiments}

Given the limitations of frontier LMs on long-horizon planning reported in prior work \citep{shojaee2025illusion,valmeekam2023planning,kambhampati2024position,valmeekam2025systematic}, our main focus is to evaluate whether \abbname{} enables reliable planning with LMs. We study the following key questions: \emph{(1) Can \abbname{} reliably solve the planning tasks considered in this work?} \emph{(2) Can \abbname{} generalize to unseen problem instances within the same domains on which it was trained?} \emph{(3) How does the SCW improve efficiency and reduce computational cost?} To answer these questions, we design a set of experiments across the planning domains considered in this work. We use a decoder-only transformer as the LM within the \abbname{} architecture. The model is trained from scratch in an autoregressive manner using the standard next-token prediction loss over all target tokens and uses greedy decoding during inference. We use two different approaches to construct the training datasets for the planning domains. For TOH, we generate all prompt--target pairs for each value of $n$ disks. During training, the model is exposed only to a random subset of these pairs, while the remaining pairs are held out for evaluation. For the Pancake Puzzle and BlocksWorld, we randomly sample a fixed number of instances for each value of $n$ and generate the complete set of prompt--target pairs associated with each instance for training. At test time, we use newly sampled problem instances for each value of $n$ that are disjoint from those used during training. More details and the hyperparameters used for each domain are provided in Appendix~\ref{sec: app-experimental_setup}.

\subsection{Accuracy Under Increasing Complexity}

 \paragraph{BlocksWorld and the Pancake Puzzle.} A key requirement of a planner is to maintain its performance as the complexity of the problem increases. To evaluate this property, we test \abbname{} across different complexity levels in each environment. For BlocksWorld and Pancake Puzzle, we generate 500 uniformly random configurations for each value of $n$ from 5 to 40 for training. Each configuration consists of a random start state; for BlocksWorld, the goal state is also sampled uniformly at random for each instance. During testing, we evaluate the model on 50 randomly sampled configurations for each $n$. The model achieves 100\% success across all test instances in both environments. A natural question is whether 50 test configurations are sufficient to represent performance at each complexity level. To answer this, we perform an exhaustive evaluation at $n=8$ over all $8! = 40{,}320$ possible start configurations. In BlocksWorld, \abbname{} successfully solves 100\% of test configurations. In the Pancake Puzzle, it fails in only one case, achieving a success rate of $99.99\%$. This failure occurs when the pancakes are already sorted, in which case the correct behavior is to generate no action. Since the LM never observes an already sorted configuration during training, it treats this case as an unsorted instance and attempts to identify the largest unsorted pancake. Consequently, it generates two flips of the smallest pancake before terminating the plan. These actions leave the state unchanged, and the goal state is still reached. Therefore, this case does not reflect a failure of the planning policy itself, but rather the generation of a plan containing unnecessary actions.

\paragraph{Extended BlocksWorld.} We consider a more general variant of BlocksWorld. In the standard setting, the initial configuration consists of a single stack of blocks. We extend this to allow multiple stacks in the initial state, introducing a broader set of combinatorial configurations, and refer to this setting as \emph{Extended BlocksWorld}. A complete example illustrating how the context window evolves with all prompts and LM outputs is provided in Appendix~\ref{sec: extended-blocksworld-prompt-example}. Similar to the standard BlocksWorld setting, we use 500 randomly sampled start--goal configuration pairs for each value of $n$ from $5$ to $40$ for training. At test time, we evaluate the model on 50 randomly sampled configurations per $n$. In this extended setting, it is not possible to cover all stack configurations during training, and the model does not observe all possible numbers of stacks for each value of $n$. Despite this, \abbname{} successfully solves all test configurations for every value of $n$ except two configurations for $n=35$ and $n=36$, achieving a success rate of $99.89\%$. This result demonstrates that the model learns the underlying planning strategy and generalizes across different stack configurations, regardless of how the blocks are distributed among the stacks.

\begin{wrapfigure}{r}{0.48\linewidth}
    \centering
    \includegraphics[width=\linewidth]{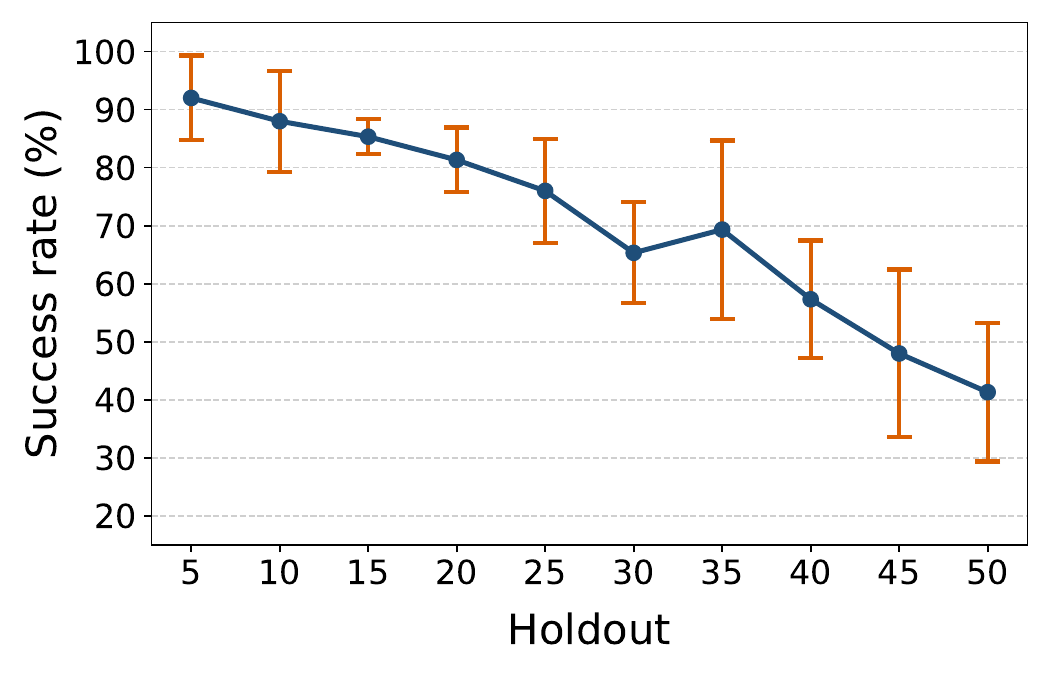}
    \caption{Mean success rate on TOH across random hold-out percentages of the training data, with error bars showing $\pm$ one standard deviation over five random seeds.}
    \label{fig:toh-random-holdout}
\end{wrapfigure}

\paragraph{Tower of Hanoi.} For TOH, we train the model on prompt--target pairs for $n=1$ to $n=15$ disks, holding out a fraction of the pairs uniformly at random to evaluate generalization. We gradually increase the hold-out percentage from $5\%$ to $50\%$. For each hold-out level, we use a distinct set of five random seeds so that the held-out pairs differ as much as possible across runs. The success rates are shown in Figure~\ref{fig:toh-random-holdout}, while the detailed results, including the values of $n$ for which the planner fails and the corresponding failure steps for one 
representative run from each hold-out setting, are provided in Appendix~\ref{sec: complete-training-instances}. As expected, the average success rate decreases as the holdout percentage increases, dropping from 92\% at a 5\% holdout rate to 41\% at 50\%. Nevertheless, even with $35\%$ of the data held out, \abbname{} achieves an average success rate of $70\%$, substantially outperforming recent methods \citep{shojaee2025illusion,malach2026infinity}, which employ much larger pretrained reasoning LMs. Also, a closer inspection of the failure cases reveals that most failures occur when the planner encounters unseen disk indices or unseen instruction numbers. In these cases, the failures are typically caused by incorrect arithmetic operations when updating the pointer or directing it to the next instruction. Therefore, we conducted an experiment in which the arithmetic operations were separated from the planning operations. The results show that the planning success rate remains at $100\%$ across all values of $n$, while requiring substantially less training data than the original setting. Moreover, the amount of training data grows only linearly with $n$, whereas the solution length grows exponentially. The results, along with a detailed description of the experiment, are provided in Section~\ref{sec: isolating arithmetic in TOH}.


\begin{table}[htbp]
\centering
\caption{Success rates (\%) across selected complexity levels
and numbers of training instances.}
\label{tab:selected-success-rates}
\setlength{\tabcolsep}{3.5pt}
\begin{tabular}{>{\centering\arraybackslash}m{2.9cm} c c c c c c c c c}
\toprule
Environment & Training Instances & \multicolumn{8}{c}{$n$} \\
\cmidrule(lr){2-2}\cmidrule(lr){3-10}
& & 6 & 10 & 15 & 20 & 25 & 30 & 35 & 40 \\
\midrule
\multirow{5}{*}{BlocksWorld}
& 500 & 100 & 100 & 100 & 100 & 100 & 100 & 100 & 100 \\
& 300 & 100 & 100 & 100 & 100 & 100 & 100 & 100 & 100 \\
& 100 & 100 & 100 & 100 & 100 & 100 & 100 & 100 & 100 \\
& 30  & 100 & 100 & 100 & 100 & 100 & 100 & 100 & 92 \\
& 5   & 94 & 80 & 100 & 76 & 78 & 54 & 30 & 0 \\
\midrule
\multirow{5}{*}{Extended BlocksWorld}
& 500 & 100 & 100 & 100 & 100 & 100 & 100 & 98 & 100 \\
& 300 & 100 & 100 & 100 & 100 & 100 & 100 & 100 & 98 \\
& 100 & 100 & 100 & 100 & 100 & 100 & 100 & 98 & 98 \\
& 30  & 100 & 100 & 100 & 100 & 100 & 98 & 100 & 94 \\
& 5   & 94 & 88 & 78 & 72 & 52 & 38 & 16 & 2 \\
\midrule
\multirow{5}{*}{Pancake}
& 500 & 100 & 100 & 100 & 100 & 100 & 100 & 100 & 100 \\
& 300 & 100 & 100 & 100 & 100 & 100 & 100 & 100 & 100 \\
& 100 & 100 & 100 & 100 & 100 & 100 & 100 & 100 & 100 \\
& 30  & 98 & 96 & 100 & 100 & 100 & 98 & 100 & 82 \\
& 5   & 78 & 66 & 74 & 54 & 64 & 48 & 20 & 0 \\
\bottomrule
\end{tabular}
\end{table}


\subsection{Scaling Policy Learning and Generalization with Training Data}

In the previous section, we showed that the planner achieves high success rates across all three planning domains. We now investigate how much training data is required to maintain this performance. Starting from 500 complete training instances for each value of $n$, we progressively reduce the training set size and evaluate the resulting performance on the test set, while keeping the model architecture and training setup fixed. A summary of the results is shown in Table~\ref{tab:selected-success-rates}, while the complete results are provided in Appendix~\ref{sec: complete-training-instances}.

For BlocksWorld, no failures occur until the number of training instances is reduced to 100. A few failure cases begin to appear at higher complexity levels when using 30 training instances. As the number of training instances decreases further, the failure rate gradually increases with $n$, such that with only 5 training instances, \abbname{} successfully solves only a small number of test cases at larger values of $n$. In Extended BlocksWorld, we observe a similar trend, although failures appear at higher training set sizes compared to the standard setting. This behavior is expected, since generalizing across different numbers of stacks requires more training configurations than the standard single-stack setting. Nevertheless, even with only 30 training instances, COC achieves higher success rates in both BlocksWorld variants than those reported for much larger pretrained reasoning LMs in previous work \citep{shojaee2025illusion,valmeekam2023planning,katz2024thought}. For the Pancake Puzzle, the planner exhibits almost no failures down to 100 training instances. Below this threshold, the number of failures increases steadily as the amount of training data decreases. With only 5 training instances, the planner fails on most test cases across all values of $n$. This stronger degradation compared to BlocksWorld is expected, since the Pancake Puzzle additionally requires the model to learn reversal operations and the corresponding state transformations.





%% file: sections/discussion.tex
\section{Discussion and Extensions}

When examining the results from separating arithmetic operations from planning operations in Section~\ref{sec: isolating arithmetic in TOH}, we observe that the pointer mechanism appears to be the component preventing the planner from achieving perfect success across all TOH instances. This raises a natural question: are pointers necessary for TOH at all? In TOH, the planner only needs to access the instruction at the top of the SCW at each step and append new instructions to the top of the SCW. At no point does it need to access an arbitrary location in the SCW. In other words, the planner only requires \emph{push} and \emph{pop} operations to interact with the SCW. This structure resembles a \emph{deterministic pushdown automaton (PDA)} \citep{hopcroft2001introduction}, where a finite-state machine is equipped with a LIFO external memory and can push to or pop from the stack.

The formulation of the TOH planner as a PDA, together with a complete example of the corresponding prompt--target pairs, is provided in Appendix~\ref{sec: TOH as PDA}. We train a new model on prompt--target pairs constructed using the PDA structure for TOH, following the same training setup as in the previous experiments. In this experiment, the dataset is generated for problem instances with $n=1$ to $n=20$ disks. Among all generated prompt--target pairs, only unique pairs are retained, while duplicate instructions are removed. Then, $15\%$ of the resulting dataset is selected uniformly at random and held out for testing. At test time, the planner solves 100\% of problem instances for $n=1$ to $n=20$, providing further evidence that the underlying planning policy is successfully learned by the LM. The PDA formulation can also be applied to other domains, as they require the same stack access operations. Since plan lengths do not grow exponentially in those domains, pointer operations are less challenging than in TOH, and we therefore do not repeat the experiments.

\paragraph{Depth-first search Formulation as a PDA.}

Thus far, we have shown that a planning algorithm can be learned specifically for each planning domain considered in our experiments. What remains to be investigated is whether a general planning algorithm can be formulated for use with COC and the SCW. One of the most widely used search algorithms for planning is depth-first search (DFS) \citep{tarjan1972depth}. DFS requires a stack and a world model, $\mathcal{T}(s' \mid s,a)$, capable of generating the child nodes of each node in the search tree. DFS can therefore be formulated as a PDA, with an LM implementing its finite-state control (see Section~\ref{sec: pda-dfs} for the detailed formulation).

We consider perfect mazes (where the paths form a tree) of sizes $11\times11$, $21\times21$, and $31\times31$\footnotemark{}\footnotetext{The maze generator requires odd dimensions when using a corridor width of one cell to ensure that the maze is bounded by walls on all sides.} and generate multiple mazes for each size. The training dataset consists of the push operations encountered during DFS searches over random start--goal pairs in these mazes. At each DFS iteration, the LM generates all push operations in a single output, with each push command followed by the complete maze representation of the corresponding valid child. We train a single LM on the prompt--target pairs from all maze sizes using a setup similar to that of the previous experiments. A complete example of the corresponding prompt--target pairs, along with the detailed experimental setup, is provided in Section~\ref{sec: prompt-target for DFS-LM}. For testing, we use a separate set of maze environments at each size to evaluate generalization to unseen mazes. At test time, the trained LM serves as the finite-state control of PDA. The results are presented in Table~\ref{tab:dfs-test-results}. The planner fails on only one test case from a $31\times31$ maze and successfully solves all other test cases.

\begin{table}[htbp]
\centering
\caption{Performance of the LM-equipped PDA on DFS across different maze sizes.} 
\label{tab:dfs-test-results}
\begin{tabular}{ccccc}
\toprule
\textbf{Maze size} &
\textbf{Mazes} &
\textbf{DFS searches} &
\textbf{Nodes generated per search} &
\textbf{Accuracy} \\
& & & \textbf{(mean $\pm$ SD)} & \textbf{(mean $\pm$ SD)} \\
\midrule
$11 \times 11$ & 15 & 150 & $25.85 \pm 4.27$    & $(100.00 \pm 0.00)\%$ \\
$21 \times 21$ & 15 & 150 & $107.32 \pm 20.73$ & $(100.00 \pm 0.00)\%$ \\
$31 \times 31$ & 15 & 150 & $242.05 \pm 37.35$ & $(99.33 \pm 2.58)\%$ \\
\bottomrule
\end{tabular}
\end{table}

To assess the performance of the LM-equipped PDA, we compare it against Qwen3.5-9B \citep{team2026qwen3} as a baseline under three prompting settings. In the first two settings, Qwen3.5-9B replaces our trained LM as the finite-state control of the same PDA and is prompted in either a zero-shot or one-shot setting. In the third setting, we evaluate Qwen3.5-9B using standard autoregressive planning, where the context accumulates throughout the planning process and the model is instructed to generate its response based only on the most recent state in the context. An example of each prompting setting is provided in Section~\ref{sec: prompt-target for DFS-Qwen}. In all settings, we use greedy decoding and evaluate the model both with and without \emph{thinking} enabled. Despite being approximately $778\times$ larger than our LM, Qwen3.5-9B performs poorly across all three settings and fails to successfully complete any test case. In most cases, the model generates an incorrect output at the first step of the search. Enabling \emph{thinking} does not resolve these errors; instead, the model often exhibits \emph{overthinking} \citep{chen2025donot,qu2025optimizing}, generating lengthy reasoning traces that attempt to correct earlier mistakes but introduce additional errors in the process.

%% file: sections/conclusion.tex
\section{Conclusion and Future Work}

In this work, we investigated the gap between the theoretical computational power of transformer-based LMs and their empirical performance on planning tasks. We proposed the \name{} (\abbname{}) architecture, in which an LM operates inside an iterative loop and interacts with a structured context window (SCW) through a pointer mechanism. Theoretically, we compared the SCW with the accumulated context used in standard full-context planning by analyzing how the prediction error evolves as the number of planning steps increases. We evaluated the proposed approach on three planning domains: Tower of Hanoi (TOH), BlocksWorld, and the Pancake Puzzle. In BlocksWorld and the Pancake Puzzle, we found that relatively small LMs trained from scratch can learn planning policies and generalize from a small number of training instances, achieving near-perfect success rates. In TOH, \abbname{} generalized to unseen instructions and achieved an average success rate of $92\%$. By separating planning from arithmetic in TOH and handling the arithmetic symbolically, we showed that perfect performance can be achieved with substantially less training data. We further showed that \abbname{} can be formulated as a deterministic pushdown automaton (PDA), eliminating the need for explicit pointer operations. Under this formulation, the planner achieved a perfect success rate on all TOH instances with up to 20 disks and successfully executed depth-first search (DFS) in perfect maze environments of varying sizes. Several directions remain for future work. First, we aim to investigate whether \abbname{} can accurately execute more sophisticated planning algorithms. Second, we plan to study whether pretrained LMs can improve length generalization and out-of-distribution generalization within the \abbname{} architecture.

%% file: sections/appendix.tex
\newpage
\begin{center}
	{\fontsize{18pt}{0pt}\selectfont \bf Appendix}
\end{center}

\startcontents            
\setcounter{tocdepth}{2}  
\printcontents{}{1}{\section*{Contents}}

\input{sections/related_work}

\section{Pseudocode for \abbname{} Architecture} \label{sec: coc pseudocode}

Algorithm~\ref{alg:computational_architecture} illustrates the iterative procedure underlying the \abbname{} architecture. At the beginning of the planning procedure, the task is placed at position $0$ in the SCW, and the pointer is initialized to $0$. At each iteration, the LM receives the current instruction and generates a new instruction, the next action, and the pointer to the next instruction to be executed. The new instruction is then written to the location specified within the instruction itself. The generated action is appended to the action sequence, and the process continues until the pointer becomes $-1$, indicating that no instructions remain to be executed. Upon termination, the accumulated action sequence is returned as the solution path.

\begin{algorithm}[t]
\caption{\abbname{}: \name{} Architecture}
\label{alg:computational_architecture}
\begin{algorithmic}[1]
\Require Initial state $\sstart$, goal state $\sgoal$, language model $\LM$
\Ensure Action sequence $\plan$ that transforms $\sstart$ to $\sgoal$

\State Initialize structured context window $\memory \gets \emptyset$
\State Initialize action sequence $\plan \gets \emptyset$
\State Write the initial instruction $\textsc{Solve}(\sstart, \sgoal)$ to $\memory[0]$
\State Set pointer $\pointer \gets 0$

\While{$\pointer \neq -1$}
    \State $\ctx \gets \textsc{Read}(\memory, \pointer)$
    \Comment{Read the instruction at $\pointer$}
    
    \State $(\mathcal{I}_{\mathrm{new}}, \mathcal{A}_{\mathrm{new}}, \pointer_{\mathrm{next}}) \gets \LM(\ctx)$
    \Comment{Generate new instruction, action, and next pointer}
    
    \State $\textsc{Write}(\memory,\mathcal{I}_{\mathrm{new}})$
    \Comment{Write generated instruction to the SCW}
    
    \State $\plan \gets \plan \circ \mathcal{A}_{\mathrm{new}}$
    \Comment{Append the new action to the action sequence}
    
    \State $\pointer \gets \pointer_{\mathrm{next}}$
\EndWhile

\State \Return $\plan$
\end{algorithmic}
\end{algorithm}

\section{Experimental Setup} \label{sec: app-experimental_setup}

In this section, we describe the tokenization process and the LM structure. For each environment, we convert every instruction example from training into a prompt--target language modeling sequence. Each dataset record contains a textual prompt $x_{prompt}$ and a textual target $y$. We train a separate tokenizer for each
environment using only the concatenated strings
\[
    x_{prompt} \; || \; \texttt{\textbackslash n} \; || \; y ,
\]
where \texttt{\textbackslash n} is the prompt--target separator. The tokenizer is a Unicode, non-byte-level BPE tokenizer \citep{sennrich2016neural} with a Metaspace pre-tokenizer and decoder. The BPE vocabulary is initialized with the set of characters appearing in the corresponding environment's prompt--target text, together with the special tokens \texttt{[PAD]}, \texttt{[BOS]}, \texttt{[EOS]}, and \texttt{[UNK]}. We do not train the tokenizer on metadata fields such as problem size, step index, or configuration identifiers; only the prompt and target strings are used. After tokenizer training, each record is encoded as
\[
    x = [\texttt{BOS}] \; || \; \mathrm{tok}(x_{prompt}) \; || \;
        \mathrm{tok}(\texttt{\textbackslash n}) \; || \;
        \mathrm{tok}(y) \; || \; [\texttt{EOS}] .
\]
We store the token index at which the target portion begins. During training, the model is optimized with next-token prediction, but loss terms corresponding to prompt reconstruction and padding tokens are masked out.

\paragraph{Model architecture.}
All experiments use a decoder-only Transformer language model. Given an input sequence of token ids $(id_1,\ldots,id_n)$, the model forms the initial hidden
states by summing learned token embeddings and learned positional
embeddings:
\[
    x_i = E_{\mathrm{tok}}(id_i) + E_{\mathrm{pos}}(i),
    \qquad i = 1,\ldots,n .
\]

\begin{figure}[t]
    \centering
    \includegraphics[width=\linewidth]{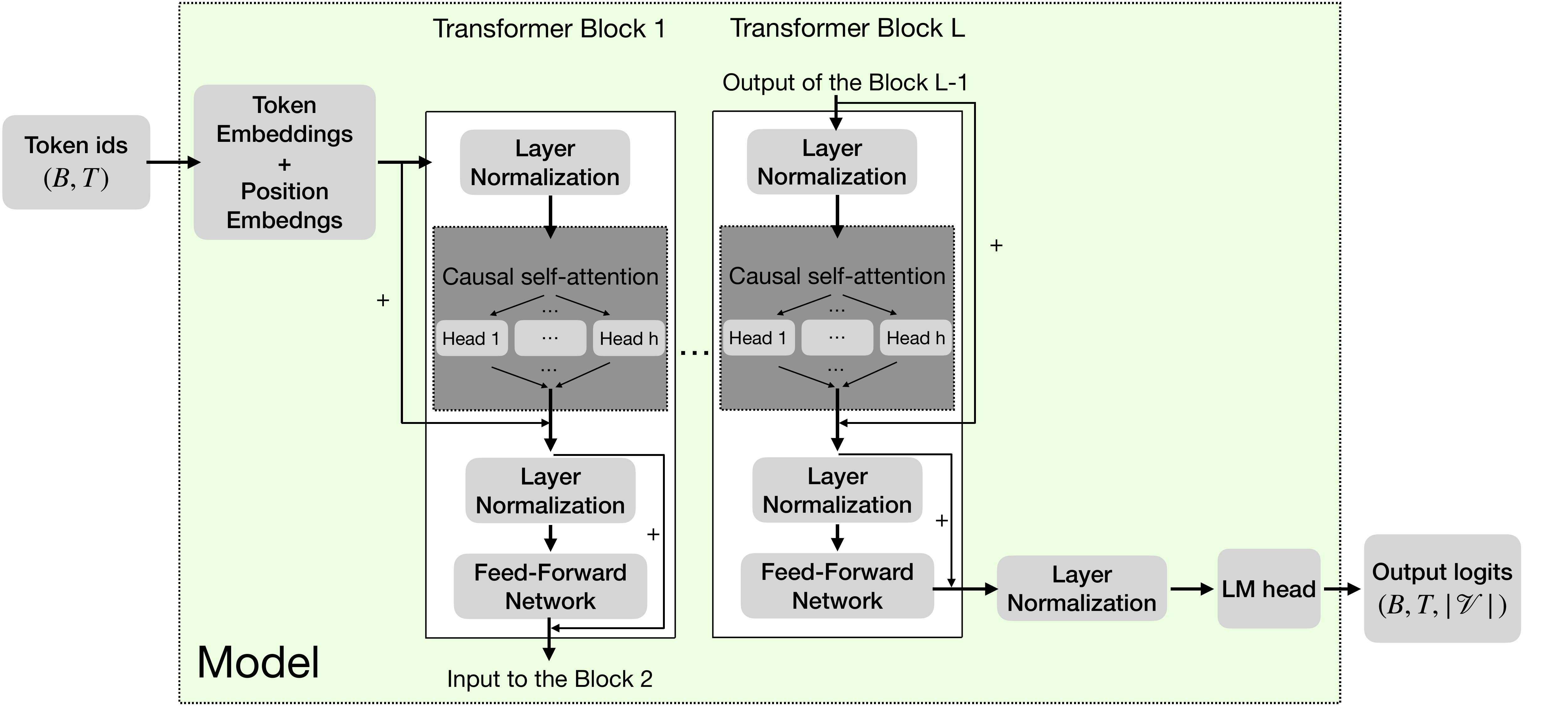}
    \caption{Decoder-only Transformer architecture used as the language model within the planner.}
    \label{fig: language model}
\end{figure}

The sequence is then processed by $L$ identical pre-normalization Transformer blocks. Each block consists of causal multi-head self-attention followed by a position-wise feed-forward network, with residual connections around both sub-layers. The attention mask is strictly causal, so position $i$ can attend only to positions $j \leq i$; padding positions are additionally masked. The feed-forward network is a two-layer MLP with a GELU nonlinearity \citep{hendrycks2016gaussian}. After the final Transformer block, we apply a final layer normalization and a linear language-modeling head to obtain logits over the tokenizer vocabulary. An overview of the model architecture is shown in Figure \ref{fig: language model}. The model hyperparameters used for each puzzle environment are reported in Table~\ref{tab: model-hyperparams}, and the corresponding training hyperparameters are reported in Table~\ref{tab: training-hyperparams}.


\begin{table}[h]
\centering
\caption{Model and tokenization hyperparameters used for each environment.}
\label{tab: model-hyperparams}
\begin{tabular}{lccc}
\toprule
\textbf{Hyperparameter} & \textbf{Tower of Hanoi} & \textbf{Blocksworld} & \textbf{Pancake} \\
\midrule
Tokenizer & Unicode BPE & Unicode BPE & Unicode BPE \\
BPE vocabulary size & 300 & 300 & 300 \\
BPE min frequency & 10 & 5 & 5 \\
BOS/EOS tokens & Yes & Yes & Yes \\
Maximum sequence length & 128 & 256 & 512 \\
Trainable parameters & 10.8 M & 25.49 M & 25.62 M \\
Hidden size $d_{\mathrm{model}}$ & 384 & 512 & 512 \\
Transformer blocks $L$ & 6 & 8 & 8 \\
Attention heads & 6 & 8 & 8 \\
Attention head dimension & 64 & 64 & 64 \\
Feed-forward dimension $d_{\mathrm{ff}}$ & 1536 & 2048 & 2048 \\
Dropout & 0.1 & 0.1 & 0.1 \\
Input/output embedding tying & Yes & Yes & Yes \\
\bottomrule
\end{tabular}
\end{table}

\begin{table}[h]
\centering
\caption{Training hyperparameters used for each environment.}
\label{tab: training-hyperparams}
\begin{tabular}{lccc}
\toprule
\textbf{Hyperparameter} & \textbf{Tower of Hanoi} & \textbf{Blocksworld} & \textbf{Pancake} \\
\midrule
Training sequence length & 128 & 256 & 512 \\
Batch size & 64 & 32 & 32 \\
Validation fraction & 0.05 & 0.05 & 0.05 \\
Training configurations per n & --- & 500 & 500\\
Epochs & 100 & 100 & 100 \\
Maximum training steps & 50{,}000 & 100{,}000 & 100{,}000 \\
Learning rate & $3 \times 10^{-4}$ & $3 \times 10^{-4}$ & $3 \times 10^{-4}$ \\
Weight decay & 0.1 & 0.1 & 0.1 \\
Gradient clipping & 1.0 & 1.0 & 1.0 \\
Warmup steps & 2{,}000 & 2{,}000 & 2{,}000 \\
Optimizer & AdamW & AdamW & AdamW \\
Learning-rate schedule & Cosine decay & Cosine decay & Cosine decay \\
Evaluation interval & 1{,}000 steps & 1{,}000 steps & 1{,}000 steps \\
Logging interval & 100 steps & 100 steps & 100 steps \\
\bottomrule
\end{tabular}
\end{table}

\section{Proofs for Theoretical Results} \label{sec: theoretical proofs}

In this section, we provide detailed proofs for the theoretical results in the paper. We first provide the theorems that we use in our derivation below.

\begin{lemma}[Lemma 2.1 in~\citep{velivckovic2025softmax}] \label{lemma: softmax dispersion in general}
Let $\mathbf{e}^{(n)} \in \mathbb{R}^{n}$ be a collection of $n$ logits going into the $\operatorname{softmax}_{\theta}$ function with temperature $\theta > 0$, bounded above and below s.t. $m \leq e_{k}^{(n)} \leq M$ for some choice of constants $m, M \in \mathbb{R}$. Then, as more items are added $(n \rightarrow +\infty)$, it must hold that, for each item $1 \leq k \leq n$,
\[
\operatorname{softmax}_{\theta}\left(\mathbf{e}^{(n)}\right)_{k}
=
\Theta\left(\frac{1}{n}\right).
\]  
\end{lemma}

\begin{lemma}[Theorem 2.2 in~\citep{velivckovic2025softmax}] \label{lemma: softmax dispersion in the network}
Let $\mathcal{X} \subset \mathbb{R}^{m}$ be a set of possible $m$-dimensional input features, and let $\mathbf{X}^{(n)} \in \mathcal{X}^{n}$ be a matrix of input features for $n$ items. Further, assume that input features come from a finite set of possible values, i.e.
$|\mathcal{X}| < |\mathbb{N}|$. Let $e_j^{(n)}=\left(\mathbf{q}^{(n)}\right)^{\top}\mathbf{k}_j^{(n)}$ where $\mathbf{q}^{(n)}=\phi(\mathbf{x}_1^{(n)},\ldots,\mathbf{x}_n^{(n)})$ and $\mathbf{K}^{(n)}=\kappa(\mathbf{x}_1^{(n)},\ldots,\mathbf{x}_n^{(n)}),$ where $\phi : \mathcal{X}^{n} \rightarrow \mathbb{R}^{k}$ and $\kappa : \mathcal{X}^{n} \rightarrow \mathbb{R}^{n\times k}$ are continuous functions, each expressible as a composition of $L$ layers $Attn_L \circ FF_L \circ \cdots \circ Attn_1 \circ FF_1$ where each layer contains a feedforward component $FF_i(\mathbf{z}_1,\ldots,\mathbf{z}_n)_k=FF_i(\mathbf{z}_k)$ or a self-attentional component $Attn_i(\mathbf{z}_1,\ldots,\mathbf{z}_n)_k=\sum_{1\leq l\leq n}\alpha_{lk}v_i(\mathbf{z}_l)$ where $\alpha_{lk}\in[0,1]$ are softmax-normalised attention coefficients and $v_i$ is a feedforward network. Then, for any $\theta>0$ and $\delta>0$, there must exist an $n\in\mathbb{N}$ such that $\operatorname{softmax}_{\theta}\left(\mathbf{e}^{(n)}\right)_k <\delta$ for all $1\leq k\leq n.$
\end{lemma}

\begin{lemma} [Proposition D.2 in \citep{dziri2023faith}] \label{thm: general probability of failure}
    Let $f_d(\mathbf{x}) = g^d(\mathbf{x})$. Assume
$\mathbb{P}\bigl(g(X)=\widehat{g}(Y)\mid X\neq Y\bigr)\leq c$
(i.e. recovering from a mistake due to the randomness of applying the estimator on an incorrect input has probability at most $c$).
If $\mathbb{P}(g\neq\widehat{g})=\epsilon>0$ with $c+\epsilon<1$, then
\[
\liminf_{d\to+\infty}
\mathbb{P}\bigl(f_d\neq\widehat{f}_d\bigr)
\geq
1-\frac{c}{c+\epsilon}.
\]
\end{lemma}

\begin{proof}[Proof of Theorem \ref{thm: main-failure-bound}]

We consider a family of planning problems $\{P_d\}_{d\geq 1}$ whose complexity is indexed by $d$. Solving a problem of complexity $d$ requires the iterative application of a transition function $g$. We denote the exact computation by $f(P_d)=g^{\,d}(P_d)$, where $g^{\,d}$ denotes the $d$-fold composition of $g$. The LM approximation of $g$ is denoted by $\widehat{g}$, thereby inducing an approximate computation $\widehat{f}(P_d)$.

At iteration $t$, let $\mathcal{W}_t$ denote the context presented to the model. We partition its tokens into a set of relevant tokens $\mathcal{R}_t$ and a set of irrelevant tokens $\mathcal{I}_t$, with $\mathcal{R}_t\cap\mathcal{I}_t=\varnothing$. Let $r_t=|\mathcal{R}_t|$ and $i_t=|\mathcal{I}_t|$ denote the numbers of relevant and irrelevant tokens, respectively. The total context length is therefore $w_t=r_t+i_t$. Throughout the analysis, we assume that the context can grow without bound and that no truncation is applied. We further assume that the amount of relevant information required at each iteration remains unchanged; hence, $r_t=r$ is constant.

Under the SCW with CoC setting, the model is provided only with the information required for the current application of $g$. In particular, the irrelevant tokens $\mathcal{I}_t$ are excluded from the context, and the model receives only the relevant tokens $\mathcal{R}_t$, so that $w_t=r$ at every iteration. Accordingly, the one-step error probability $\epsilon=\mathbb{P}(g\neq\widehat{g})$ and the recovery bound $c$, defined by $\mathbb{P}\!\left(g(X)=\widehat{g}(Y)\mid X\neq Y\right)\leq c$, are treated as iteration-independent constants. Consequently, Lemma~\ref{thm: general probability of failure} applies directly to the SCW setting and yields
\[
\liminf_{d\to+\infty}
\mathbb{P}\!\left(f(P_d)\neq\widehat{f}(P_d)\right)
\geq
1-\frac{c}{c+\epsilon}.
\]

The situation differs when the entire accumulated context is provided to the model at every iteration. In this case, the context generated and used in previous iterations contributes irrelevant information to the current application of $g$. We model this accumulation by $w_t=r+t\rho$, where $\rho$ is a constant denoting the number of irrelevant tokens added per planning iteration. Thus, while the amount of relevant information remains fixed, both the context length and the proportion of irrelevant tokens increase with $t$. A consequence of this growing context under softmax attention is \emph{softmax dispersion}, as characterized in Lemmas~\ref{lemma: softmax dispersion in general} and~\ref{lemma: softmax dispersion in the network}. In particular, Lemma~\ref{lemma: softmax dispersion in general} shows that, when the attention logits remain uniformly bounded, each individual attention coefficient scales as $\Theta(1/n)$ as the number of tokens $n$ grows. Lemma~\ref{lemma: softmax dispersion in the network} extends this behavior to the Transformer architecture, showing that all individual attention coefficients become arbitrarily small as the context length increases. 

Motivated by these results, under the full-context setting we allow $\epsilon$ to vary across iterations and denote it by $\epsilon_t$. Similar to \citet{dziri2023faith}, we assume that $g$ and $\widehat{g}$ have a low collision rate, so that accidental recovery from an incorrect response is unlikely. 
For simplicity, we treat the recovery bound as approximately iteration-independent, i.e., $c_t \approx c$ for all $t$. In the following, we focus on how the accumulation of irrelevant context affects $\epsilon_t$.

Our notation for the Transformer architecture closely follows \citet{barbero2024transformers}. We consider a depth-$L$ decoder-only Transformer with residual connections and MLP layers. For an input sequence of $n$ tokens, let $\mathbf{q}_i^{(\ell)}$, $\mathbf{k}_i^{(\ell)}$, and $\mathbf{v}_i^{(\ell)} \in \mathbb{R}^{d_{\mathrm{model}}}$ denote the query, key, and token representations of token $i$ at layer $\ell$, respectively. The initial representation is given by the token embedding $\mathbf{v}_i^{(0)}=x_i$, and at each layer $\ell$ the query and key representations are obtained from the normalized token representation as $\mathbf{q}_i^{(\ell)}=W_Q^{(\ell)} \operatorname{norm}_1^{(\ell)}(\mathbf{v}_i^{(\ell)})$ and
$\mathbf{k}_i^{(\ell)}=W_K^{(\ell)} \operatorname{norm}_1^{(\ell)}(\mathbf{v}_i^{(\ell)})$, respectively. We further denote by $\mathbf{p}_{i,j} \in \mathbb{R}^{2e}$ the positional-encoding information between tokens $i$ and $j$. The attention coefficients at layer $\ell$ are given by

\begin{equation*}
\alpha_{i,j}^{(\ell)}
=
\frac{
\exp\left(
\kappa\left(
\mathbf{q}_i^{(\ell)},
\mathbf{k}_j^{(\ell)},
\mathbf{p}_{i,j}
\right)
\right)
}{
\sum_{w=1}^{i}
\exp\left(
\kappa\left(
\mathbf{q}_i^{(\ell)},
\mathbf{k}_w^{(\ell)},
\mathbf{p}_{i,w}
\right)
\right)
},
\qquad j \leq i,
\end{equation*}

with $\alpha_{i,j}^{(\ell)}=0$ for $j>i$, where
$\kappa:\mathbb{R}^{d_{\mathrm{model}}}\times\mathbb{R}^{d_{\mathrm{model}}}\times\mathbb{R}^{2e}\rightarrow\mathbb{R}$
is a scalar attention-logit function that maps the query, key, and positional
information to an attention score. For token $i$, the output of the self-attention sublayer together with the residual connection is

\begin{equation*}
\mathbf{z}_i^{(\ell)}
=
\sum_{j=1}^{i}
\alpha_{i,j}^{(\ell)}
\operatorname{norm}_1^{(\ell)}
\left(\mathbf{v}_j^{(\ell)}\right)
+
\mathbf{v}_i^{(\ell)}.
\end{equation*}

The MLP sublayer then updates the token representation as

\begin{equation*}
\mathbf{v}_i^{(\ell+1)}
=
\psi^{(\ell)}
\left(
\operatorname{norm}_2^{(\ell)}
\left(\mathbf{z}_i^{(\ell)}\right)
\right)
+
\mathbf{z}_i^{(\ell)}.
\end{equation*}

Here, $\psi^{(\ell)}$ denotes the MLP at layer $\ell$, and
$\operatorname{norm}_1^{(\ell)}$ and $\operatorname{norm}_2^{(\ell)}$
denote the corresponding normalization operations. Since next-token prediction is based on the representation of the last input token, we focus on position $n$. At layer $\ell$, its representation is given by

\begin{equation*}
\mathbf{z}_n^{(\ell)}
=
\sum_{j=1}^{n}
\alpha_{n,j}^{(\ell)}
\operatorname{norm}_1^{(\ell)}
\left(\mathbf{v}_j^{(\ell)}\right)
+
\mathbf{v}_n^{(\ell)},
\end{equation*}

followed by

\begin{equation*}
\mathbf{v}_n^{(\ell+1)}
=
\psi^{(\ell)}
\left(
\operatorname{norm}_2^{(\ell)}
\left(\mathbf{z}_n^{(\ell)}\right)
\right)
+
\mathbf{z}_n^{(\ell)}.
\end{equation*}

After applying these updates through all $L$ Transformer layers, the final
representation of the last token is

\begin{equation*}
\mathbf{y}_n
=
\operatorname{norm}_3
\left(\mathbf{v}_n^{(L)}\right).
\end{equation*}

This final representation $\mathbf{y}_n$ is subsequently used to compute the logits for predicting the next token. We consider greedy decoding, as commonly used in deterministic planning and search algorithms, and denote the correct next token by \(\tau^\star\). We consider a Transformer consisting of a single self-attention layer and make the following additional assumptions.

\begin{assumption} \label{assumption: Relevant-irrelevant attention-logit separation}

For the query corresponding to the final token, the pre-softmax attention logits satisfy

$$
s_{n,j}
=
\begin{cases}
M, & j\in\mathcal{R},\\[2mm]
m, & j\in\mathcal{I},
\end{cases}
$$

where $\Delta:=M-m>0$. The quantities $M$ and $m$ are finite and independent of the iteration.
\end{assumption}

\begin{assumption} \label{assumption: Layer-normalization approximation}
Each normalization operation is approximated by a fixed positive scaling factor: 
\[
\operatorname{norm}_i(\mathbf{x}) \approx \beta_i\mathbf{x},\ \ \ \beta_i>0
\]
\end{assumption}

\begin{assumption} \label{assumption: Locally affine ReLU MLP}
We assume that the MLP is ReLU-based and that the representations $\mathbf{z}_t$ remain within the same ReLU activation region for all $t\geq0$. Consequently, the MLP admits a common affine representation across iterations:
\[
\psi(\beta_2\mathbf{z}_t)=\mathbf{A}\mathbf{z}_t+\mathbf{c},
\qquad t\geq 0.
\]
\end{assumption}

Consider the first iteration, where the context contains only the $r$ relevant tokens. Since every relevant token has attention logit \(M\),

$$
\alpha_{n,j}^{(0)}
=
\frac{e^M}{r e^M}
=
\frac{1}{r}.
$$

Define the average relevant-token representation as

$$
\overline{\mathbf{v}}_R
=
\frac{1}{r}
\sum_{j\in\mathcal{R}}\mathbf{v}_j.
$$

Using Assumption \ref{assumption: Layer-normalization approximation}, the representation following the attention sublayer is $\mathbf{z}_{n}^{(0)}[0]
=\mathbf{v}_n+\frac{\beta_1}{r}\sum_{j\in\mathcal{R}}\mathbf{v}_j$, and therefore $\mathbf{z}_{n}^{(0)}[0] = \mathbf{v}_n + \beta_1\overline{\mathbf{v}}_R$. Using Assumption \ref{assumption: Locally affine ReLU MLP}, $\mathbf{v}_{n}^{(1)}[0]=\psi(\beta_2\mathbf{z}_{n}^{(0)}[0])+\mathbf{z}_{n}^{(0)}[0]=(\mathbf{I}+\mathbf{A})\mathbf{z}_{n}^{(0)}[0]+\mathbf{c}$. After the final normalization, $\mathbf{y}_n[0]=\beta_3\left[(\mathbf{I}+\mathbf{A})\mathbf{z}_{n}^{(0)}[0]+\mathbf{c}\right]$. Hence,

\begin{equation}
\boxed{
\mathbf{y}_n[0]
=
\beta_3
\left[
(\mathbf{I}+\mathbf{A})
\left(
\mathbf{v}_n+\beta_1\overline{\mathbf{v}}_R
\right)
+\mathbf{c}
\right]
}
\label{eq:y-clean}
\end{equation}

The corresponding vocabulary logits are $\boldsymbol{\ell}[0]=\mathbf{W}_U\mathbf{y}_n[0]+\mathbf{b}_U$. Define $\mathbf{W}:=\beta_3\mathbf{W}_U(\mathbf{I}+\mathbf{A})$ and $\mathbf{b}:=\beta_3\mathbf{W}_U\mathbf{c}+\mathbf{b}_U$. Then

\begin{equation}
\boxed{
\boldsymbol{\ell}[0]
=
\mathbf{W}
\left(
\mathbf{v}_n+\beta_1\overline{\mathbf{v}}_R
\right)
+
\mathbf{b}
}.
\label{eq:logits-clean}
\end{equation}

The next-token prediction at iteration $t=0$ is then
\begin{equation}
\widehat{\tau}_0
=
\arg\max_k \ell_{0,k},
\end{equation}
and its failure probability is assumed to be
\begin{equation}
\Pr(\widehat{\tau}_0 \neq \tau^\star) = \epsilon.
\label{eq:base-error}
\end{equation}

At iteration $t$, the attention softmax denominator is $Z_t=r e^M+\rho t\,e^m$. Using Assumption \ref{assumption: Relevant-irrelevant attention-logit separation}, $Z_t=e^m\left(r e^\Delta+\rho t\right)$. Thus, the attention weight assigned to each relevant token is $\alpha_R(t)=
\frac{e^\Delta}{r e^\Delta+\rho t}$, whereas each irrelevant token receives $\alpha_I(t)=\frac{1}{r e^\Delta+\rho t}$. Define $ \overline{\mathbf{v}}_{I,t}=\frac{1}{\rho t}\sum_{j\in\mathcal{I}_t}\mathbf{v}_j$. The representation following the attention sublayer is

\begin{equation}
\boxed{
\begin{aligned}
\mathbf{z}_{n}^{(0)}[t]
&=
\mathbf{v}_n
+
\frac{\beta_1}{r e^\Delta+\rho t}
\left[
e^\Delta \sum_{j\in\mathcal{R}}\mathbf{v}_j
+
\sum_{j\in\mathcal{I}_t}\mathbf{v}_j
\right] \\
&=
\mathbf{v}_n
+
\beta_1
\left[
\frac{r e^\Delta}{r e^\Delta+\rho t}
\overline{\mathbf{v}}_R
+
\frac{\rho t}{r e^\Delta+\rho t}
\overline{\mathbf{v}}_{I,t}
\right].
\end{aligned}
}
\label{eq:z-perturbed}
\end{equation}

The representation following the MLP and residual connection is $\mathbf{v}_n^{(1)}[t]=(\mathbf{I}+\mathbf{A})\mathbf{z}_{n}^{(0)}[t]+\mathbf{c}$, and hence $\mathbf{y}_{n}[t]=\beta_3\left[(\mathbf{I}+\mathbf{A})\mathbf{z}_{n}^{(0)}[t]+\mathbf{c}\right]$. The corresponding vocabulary logits are

\begin{equation}
\boxed{\boldsymbol{\ell}[t]=\mathbf{W}\left[\mathbf{v}_n+\beta_1\left(\frac{r e^\Delta}{r e^\Delta+\rho t}\overline{\mathbf{v}}_R+
\frac{\rho t}{r e^\Delta+\rho t}\overline{\mathbf{v}}_{I,t}\right)\right]+\mathbf{b}.}
\label{eq:logits-perturbed}
\end{equation}

Subtracting \eqref{eq:logits-clean} from \eqref{eq:logits-perturbed}, and noting that the irrelevant tokens are prepended to the context so that $\mathbf{v}_n$ is identical in both equations, gives
\begin{equation}
\boxed{\boldsymbol{\ell}[t]-\boldsymbol{\ell}[0]=\frac{\beta_1\rho t}{r e^\Delta+\rho t}\mathbf{W}\left(\overline{\mathbf{v}}_{I,d}
-\overline{\mathbf{v}}_R\right).}
\label{eq:logit-change}
\end{equation}

Thus, the effect of the inserted irrelevant tokens on greedy decoding is entirely determined by their induced change in the vocabulary-logit vector.

\paragraph{Pairwise Failure Probability}

To express $\epsilon_t$ in terms of $\epsilon$, an additional stochastic assumption is required. Consider the correct token $\tau^\star$ and a dominant incorrect competitor $\tau_c$. Define their logit margin by $G_t=\ell_{\tau^\star}[t]-\ell_{\tau_c}[t]$. Let $\mathbf{w}^{\top}=\mathbf{W}_{\tau^\star,:}-\mathbf{W}_{\tau_c,:}$. We make the following additional assumption.

\begin{assumption}
\label{assumption: stochastic-margin}
For relevant and irrelevant tokens, define the projected contributions
\[
X_j=\mathbf{w}^{\top}\mathbf{v}_j,
\quad j\in\mathcal{R},
\qquad
U_j=\mathbf{w}^{\top}\mathbf{v}_j,
\quad j\in\mathcal{I}_t,
\]
and assume
\[
X_j \overset{\mathrm{i.i.d.}}{\sim} \mathcal{N}(\mu,\sigma^2),
\qquad
U_j \overset{\mathrm{i.i.d.}}{\sim} \mathcal{N}(0,\sigma^2),
\qquad
\mu>0,
\]
with the relevant and irrelevant contributions mutually independent. We further assume that the context-independent contribution to the $\tau^\star$-versus-$\tau_c$ margin vanishes,
\[
\mathbf{w}^{\top}\mathbf{v}_n+b_{\tau^\star}-b_{\tau_c}=0.
\]
\end{assumption}

Under Assumption~\ref{assumption: stochastic-margin}, the clean pairwise margin is proportional to $S_0=\sum_{j=1}^{r}X_j$. Since $S_0\sim\mathcal{N}
\left(r\mu,r\sigma^2\right)$, the clean pairwise failure probability is
\[
\epsilon
=
\Pr(S_0\leq0)
=
\Phi
\left(
-\frac{\sqrt{r}\mu}{\sigma}
\right),
\]
where $\Phi$ denotes the cumulative distribution function (CDF) of the standard normal distribution. Therefore,
\begin{equation}
\boxed{
\Phi^{-1}(\epsilon)
=
-\frac{\sqrt{r}\mu}{\sigma}
}
\label{eq:epsilon-calibration}
\end{equation}

At iteration $t$, all common positive factors can be removed when determining the sign of the pairwise margin. The relevant random variable is consequently $S_t=e^\Delta\sum_{j=1}^{r}X_j+\sum_{j=1}^{\rho t}U_j$. Its mean is $\mathbb{E}[S_t]=r e^\Delta\mu$, and its variance is $\operatorname{Var}(S_t)=\sigma^2\left(r e^{2\Delta}+\rho t\right)$. Hence,

$$
S_t
\sim
\mathcal{N}
\left(
r e^\Delta\mu,\,
\sigma^2
\left(
r e^{2\Delta}+\rho t
\right)
\right).
$$

The pairwise failure probability is therefore $\epsilon_t=\Pr(S_t\leq0)$, giving

\begin{equation}
\boxed{
\begin{aligned}
\epsilon_t
&=
\Phi
\left(
-\frac{r e^\Delta\mu}
{\sigma\sqrt{r e^{2\Delta}+\rho t}}
\right)
&=
\Phi
\left[
\frac{\Phi^{-1}(\epsilon)}
{
\sqrt{
1+
\dfrac{\rho t}
{r e^{2\Delta}}
}
}
\right].
\end{aligned}
}
\label{eq:pairwise-final}
\end{equation}

\Eqref{eq:pairwise-final} satisfies $\epsilon_0=\epsilon$. For \(\epsilon<1/2\), one has $\Phi^{-1}(\epsilon)<0$, and therefore \(\epsilon_t\) increases monotonically with \(t\). Under Assumption \ref{assumption: Relevant-irrelevant attention-logit separation},

$$
\frac{\rho t}{r e^{2\Delta}}
\longrightarrow\infty
\qquad
\text{as}
\qquad
t\longrightarrow\infty.
$$

Consequently, at the final iteration, where $t=d$,
\begin{equation}
\boxed{
\begin{aligned}
\lim_{d\rightarrow\infty}
\frac{\Phi^{-1}(\epsilon)}
{\sqrt{1+\dfrac{\rho d}{r e^{2\Delta}}}}
&=0,
\qquad
\lim_{d\rightarrow\infty}\epsilon_d
=
\Phi(0)
=
\frac{1}{2}.
\end{aligned}
}
\label{eq:pairwise-limit}
\end{equation}

The limit \(1/2\) corresponds to a binary comparison between the correct token and one competing token. The result above considers the correct token against a single competing token. To extend the analysis to greedy decoding over the full vocabulary, let $V$ denote the vocabulary size. We make one additional assumption for this full vocabulary case: after normalization, the vocabulary logits follow the symmetric Gaussian model
$$
L_{\tau^\star}[t]=a_t+Z_0,
\qquad
L_k[t]=Z_k,\quad k\neq \tau^\star,
$$
where
$$
Z_0,\ldots,Z_{V-1}
\overset{\mathrm{i.i.d.}}{\sim}
\mathcal N(0,1).
$$

Using the attenuation factor obtained from the pairwise analysis, $\lambda_t=\frac{1}{\sqrt{1+\dfrac{\rho t}{r e^{2\Delta}}}}$, we write
$a_t=\lambda_t a_0$, where $a_0$ denotes the correct-token advantage in the absence of irrelevant tokens. Define

$$
\mathcal C_V(a)
:=
\int_{-\infty}^{\infty}
\phi(z)\,
\Phi(z+a)^{V-1}\,dz,
$$

where $\phi$ is the standard normal density function. Then

$$
\mathcal C_V(a)
=
\Pr\left(
L_{\tau^\star}>
\max_{k\neq \tau^\star}L_k
\right).
$$

Since the clean-context failure probability is \(\epsilon\), $1-\epsilon=\mathcal C_V(a_0)$, and therefore $a_0=\mathcal C_V^{-1}(1-\epsilon)$. Hence,

$$
a_t
=
\frac{
\mathcal C_V^{-1}(1-\epsilon)
}{
\sqrt{
1+\dfrac{\rho t}{r e^{2\Delta}}
}
}.
$$

The full-vocabulary failure probability at iteration \(t\) is therefore

$$
\boxed{
\epsilon_t^{(V)}
=
1-
\mathcal C_V
\left(
\frac{
\mathcal C_V^{-1}(1-\epsilon)
}{
\sqrt{
1+\dfrac{\rho t}{r e^{2\Delta}}
}
}
\right).
}
$$

At \(t=0\), $\epsilon_0^{(V)}=\epsilon$. Since \(\Delta=M-m\) is finite under Assumption \ref{assumption: Relevant-irrelevant attention-logit separation}, and thus

$$
\lambda_t\rightarrow0
\qquad\text{as}\qquad
t\rightarrow\infty.
$$

Consequently, at the final iteration, where $t=d$,

$$
a_d\rightarrow0.
$$

In this limit, $L_{\tau^\star},L_1,\ldots,L_{V-1}\overset{\mathrm{i.i.d.}}{\sim}\mathcal{N}(0,1)$. By symmetry, each of the $V$ vocabulary tokens is therefore equally likely to attain the largest logit, giving

\begin{equation}
\lim_{d\rightarrow\infty}
\Pr(\widehat{\tau}_d=\tau^\star)
=
\frac{1}{V}.
\end{equation}

Hence,

$$
\boxed{
\lim_{d\rightarrow\infty}
\epsilon_d^{(V)}
=
1-\frac1V.
}
$$

For \(V=2\), this reduces to
$$
\lim_{d\rightarrow\infty}\epsilon_d^{(2)}
=
\frac12,
$$

which agrees with the pairwise result. Moreover, since
\[
a_t
=
\frac{a_0}{\sqrt{1+\dfrac{\rho t}{r e^{2\Delta}}}}
=
\Theta\!\left(t^{-1/2}\right),
\]
and $\mathcal C_V$ is differentiable at zero with $\mathcal C_V(0)=1/V$ and $\mathcal C_V'(0)>0$, a first-order expansion around zero gives
\[
\mathcal C_V(a_t)
=
\frac{1}{V}
+
\mathcal C_V'(0)a_t
+
o(a_t).
\]
Hence,
\[
\boxed{
1-\frac{1}{V}-\epsilon_t
=
\mathcal C_V(a_t)-\frac{1}{V}
=
\Theta\!\left(t^{-1/2}\right).
}
\]

We now incorporate the iteration-dependent one-step error probability into Lemma~\ref{thm: general probability of failure}. For $P_d$, let \(\mathcal{E}_t\) denote the event that the computation is incorrect after the \(t\)-th application of \(g\), and define

$$
p_t:=\mathbb{P}(\mathcal{E}_t),
\qquad
p_0=0.
$$

In particular,

$$
p_d
=
\mathbb{P}\!\left(
f(P_d)\neq\widehat f(P_d)
\right).
$$

At iteration \(t\), $\epsilon_t$ by definition is:

$$
\mathbb{P}(\mathcal{E}_t\mid \mathcal{E}_{t-1}^{c})=\epsilon_t.
$$

As in Lemma~\ref{thm: general probability of failure}, assume that recovery from an already incorrect state occurs with probability at most \(c\):

$$
\mathbb{P}(\mathcal{E}_t^{c}\mid \mathcal{E}_{t-1})\leq c.
$$

Therefore, $\mathbb{P}(\mathcal{E}_t\mid \mathcal{E}_{t-1})\geq 1-c$. Conditioning on whether the computation at iteration \(t-1\) is correct gives

\begin{equation}
\begin{aligned}
p_t
&=
\mathbb{P}(\mathcal{E}_t\mid \mathcal{E}_{t-1}^{c})(1-p_{t-1})
+
\mathbb{P}(\mathcal{E}_t\mid \mathcal{E}_{t-1})p_{t-1}
\\
&\geq
\epsilon_t(1-p_{t-1})
+
(1-c)p_{t-1}
\\
&=
\epsilon_t
+
(1-c-\epsilon_t)p_{t-1}.
\end{aligned}
\end{equation}

Hence,
\begin{equation}
\boxed{
p_t
\geq
\epsilon_t+
(1-c-\epsilon_t)p_{t-1}.
}
\label{eq:iteration-dependent-recursion}
\end{equation}

By the assumptions of Lemma~\ref{thm: general probability of failure}, $c+\epsilon_t<1\ \text{for all }t$. Hence, $1-c-\epsilon_t>0$, and recursively applying
\eqref{eq:iteration-dependent-recursion}, with \(p_0=0\), yields
\begin{equation}
\boxed{
p_d
\geq
\sum_{t=1}^{d}
\epsilon_t
\prod_{j=t+1}^{d}
\left(1-c-\epsilon_j\right).
}
\label{eq:finite-d-variable-epsilon}
\end{equation}

For the accumulated-context setting, the previously derived full-vocabulary one-step error probability is
\begin{equation}
\boxed{
\epsilon_t=1-\mathcal{C}_V\left(\frac{\mathcal{C}_V^{-1}(1-\epsilon)}{\sqrt{1+\dfrac{\rho t}{r e^{2\Delta}}}}\right).
}
\label{eq:epsilon-t-full-vocab}
\end{equation}

For convenience, define

$$
s_t
:=
\mathcal{C}_V
\left(
\frac{
\mathcal{C}_V^{-1}(1-\epsilon)
}{
\sqrt{
1+\dfrac{\rho t}{r e^{2\Delta}}
}
}
\right),
$$

so that $\epsilon_t=1-s_t$. Substituting \(\epsilon_t=1-s_t\) into \eqref{eq:finite-d-variable-epsilon} gives
\begin{equation}
\boxed{
p_d
\geq
\sum_{t=1}^{d}
(1-s_t)
\prod_{j=t+1}^{d}
(s_j-c).
}
\label{eq:finite-d-full-vocab-bound}
\end{equation}

Since $\Delta=M-m$ is finite,

$$
\frac{
\mathcal{C}_V^{-1}(1-\epsilon)
}{
\sqrt{
1+\dfrac{\rho t}{r e^{2\Delta}}
}
}
\longrightarrow 0
\qquad
\text{as }t\rightarrow\infty.
$$

Therefore, $s_t\longrightarrow\mathcal{C}_V(0)=\frac{1}{V}$. Equivalently, the recurrence $p_t\geq(1-s_t)+(s_t-c)p_{t-1}$ asymptotically approaches

$$
p_t
\geq
1-\frac{1}{V}
+
\left(\frac{1}{V}-c\right)p_{t-1}.
$$

Let $p_\star:=\liminf_{t\rightarrow\infty}p_t$. Taking the limit inferior gives

$$
p_\star
\geq
1-\frac{1}{V}
+
\left(\frac{1}{V}-c\right)p_\star.
$$

Thus, $\left(c+1-\frac{1}{V}\right)p_\star\geq1-\frac{1}{V}$, and therefore

\begin{equation}
\boxed{
\liminf_{d\rightarrow\infty}
\mathbb{P}\!\left(
f(P_d)\neq\widehat{f}(P_d)
\right)
\geq
\frac{
1-\frac{1}{V}
}{
c+1-\frac{1}{V}
}
=
1-
\frac{c}{
c+1-\frac{1}{V}
}
}
\label{eq:full-context-final-bound}
\end{equation}

To compare this result with Lemma~\ref{thm: general probability of failure}, recall that the iteration-independent error assumption yields

$$
\mathrm{LB}_{\mathrm{const}}
=
1-\frac{c}{c+\epsilon}
=
\frac{\epsilon}{c+\epsilon},
$$

whereas the accumulated-context setting yields

$$
\mathrm{LB}_{\mathrm{acc}}
=
1-\frac{c}{c+1-\frac{1}{V}}
=
\frac{1-\frac{1}{V}}{c+1-\frac{1}{V}}.
$$

Assuming that the model performs better than random guessing in the fixed context, $\epsilon<1-\frac{1}{V}$, and since the function $h(x)=\frac{x}{c+x}
$ is strictly increasing for \(x>0\), it follows that

$$
\boxed{
\mathrm{LB}_{\mathrm{acc}}>\mathrm{LB}_{\mathrm{const}}.
}
$$

Indeed,

$$
\mathrm{LB}_{\mathrm{acc}}-\mathrm{LB}_{\mathrm{const}}
=
\frac{
c\left(1-\frac{1}{V}-\epsilon\right)
}{
\left(c+1-\frac{1}{V}\right)(c+\epsilon)
}
>0.
$$

Hence, accounting for the growth of the one-step error caused by the accumulation of irrelevant context produces a strictly larger asymptotic lower bound on the overall failure probability than treating the one-step error as the constant clean-context value \(\epsilon\). For large vocabularies, \(1-\frac{1}{V}\approx1\), and the new bound approaches $\mathrm{LB}_{\mathrm{acc}}\approx\frac{1}{1+c}$.
\end{proof}


\section{Out of Distribution Performance} \label{sec: out-of-distribution}

While the overall performance is strong, our approach does not generalize substantially beyond the training distribution. Without pre-training, the model does not generalize to larger Pancake or BlocksWorld instances because it has not learned the meaning of tokens corresponding to larger problem sizes. More generally, it does not reliably generalize to out-of-distribution problem instances. For example, replacing a block index in BlocksWorld with an unseen index causes the planning procedure to fail. Similarly, the LM does not generalize to DFS on open maps. This is partly because DFS without a visited set can revisit duplicate states, and partly because the model has not learned a sufficiently general map model from the perfect mazes observed during training. One possible direction is to use a pretrained model with fine-tuning or few-shot learning on a small number of problem instances. However, the primary goal of this work is to demonstrate that LMs can perform long-horizon planning within the COC architecture using the SCW. Future work will investigate broader problem distributions and improved length generalization.

\section{Isolating Planning from Arithmetic in TOH} \label{sec: isolating arithmetic in TOH}

\begin{figure}[h]
    \centering
    \includegraphics[width=\linewidth]{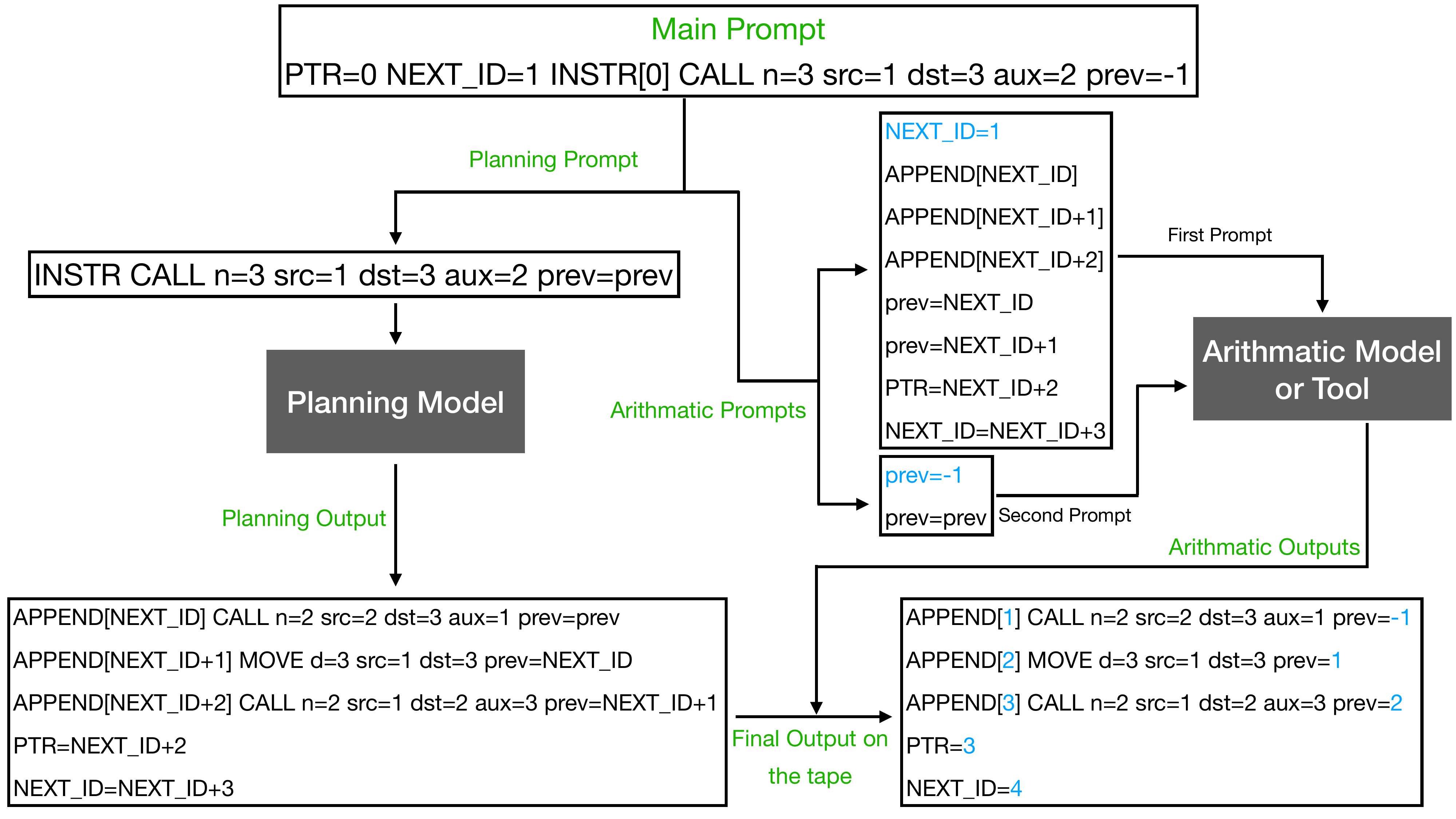}
    \caption{One iteration of the planning loop when planning and arithmetic operations are separated. The planning language model generates symbolic instructions, while the arithmetic operations are performed by a dedicated arithmetic module before the updated prompt is written back to the tape.}
    \label{fig:planning-only-structure}
\end{figure}

The results in Figure~\ref{fig:toh-random-holdout} show that holding out the same percentage of training data can lead to different levels of 
generalization depending on which prompt--target pairs are excluded from training. Motivated by this finding, we design another experiment to investigate the minimum amount of additional training data required as the number of disks increases in order to maintain a $100\%$ success rate. For this experiment, we separate the planning and arithmetic operations so that they are handled independently. The prompt is passed to the language model using symbolic placeholders, and the model generates target instructions containing symbolic arithmetic expressions rather than explicit numerical values. These symbolic expressions, along with the current values of the symbols, are then passed to an arithmetic module to obtain the updated numerical values. Figure~\ref{fig:planning-only-structure} illustrates how each iteration of the planning loop is processed, from receiving the current prompt to writing the final output on the tape. The arithmetic operations are handled separately, either by a dedicated language model trained for arithmetic reasoning or by an external tool invoked at each iteration.

\begin{wrapfigure}{r}{0.45\linewidth}
    \centering
    \includegraphics[width=\linewidth]{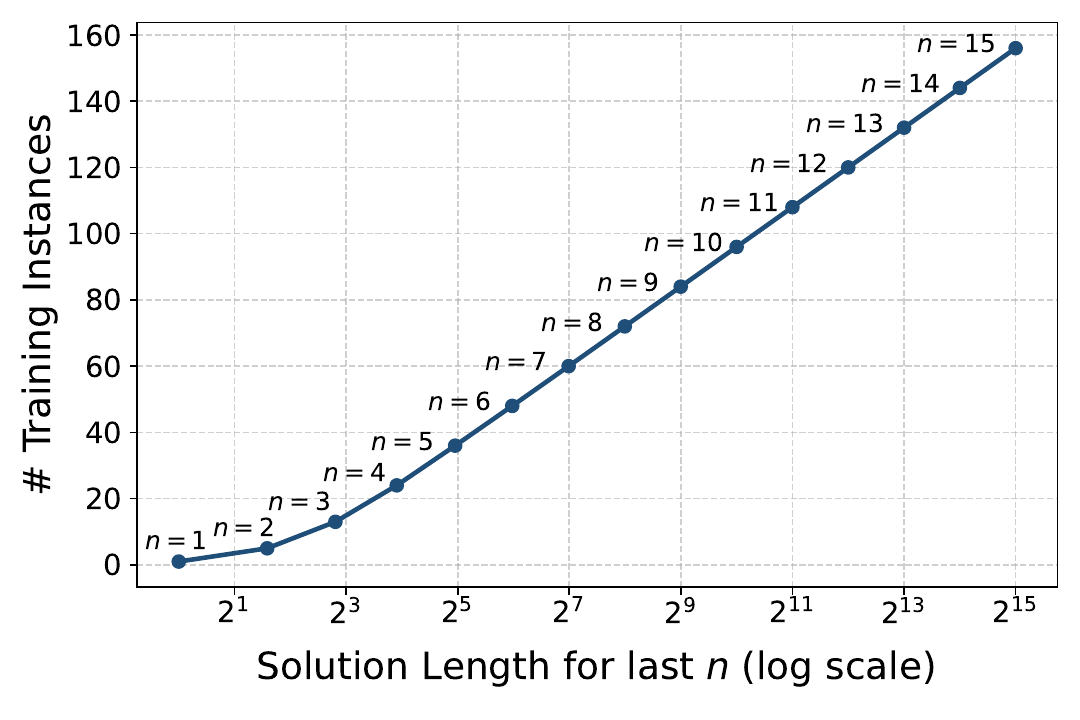}
    \caption{Planning instructions required for training to solve TOH instances up to disk count $n$. The x-axis is logarithmic.}
    \label{fig:toh-encoded-sequences-moves-second-approach}
\end{wrapfigure}

In Figure \ref{fig:toh-encoded-sequences-moves-second-approach}, we show the number of prompt--target pairs that are added to the training set as the number of disks $n$ increases in order to maintain a $100\%$ success rate for all values of $n$ up to that complexity level. In this experiment, we use the same LM architecture as in the previous experiments, while arithmetic operations are handled by a built-in module that computes the required numerical values exactly. As shown in the figure, the solution length grows exponentially with the number of disks, while the amount of additional training data required to maintain perfect performance grows only linearly. This stark contrast provides further evidence that the primary source of failures observed in the original TOH experiments is not the planning procedure itself, but rather inaccuracies in the arithmetic operations required to update the pointers. 

Below, we present the planning operations for values of $n \in [1,3]$. The structure of the prompts and targets is identical to that of the original setting; the only difference is that the arithmetic values are represented symbolically rather than being explicitly computed.

\begin{center}
\begin{tcolorbox}[
  colback=gray!12,
  colframe=gray!60,
  colbacktitle=gray!90,
  coltitle=gray!15,
  title={Example: Decomposition Operations for Tower of Hanoi with $n \in [1,3]$},
  fonttitle=\bfseries,
  sharp corners,
  boxrule=0.6pt,
  breakable
]
\small

\textbf{Operation 1}

\textbf{Prompt}
\begin{quote}
\ttfamily
INSTR CALL n=3 src=1 dst=3 aux=2 prev=prev
\end{quote}

\textbf{Target}
\begin{quote}
\ttfamily
APPEND[NEXT\_ID] CALL n=2 src=2 dst=3 aux=1 prev=prev\\
APPEND[NEXT\_ID+1] MOVE d=3 src=1 dst=3 prev=NEXT\_ID\\
APPEND[NEXT\_ID+2] CALL n=2 src=1 dst=2 aux=3 prev=NEXT\_ID+1\\
PTR=NEXT\_ID+2 NEXT\_ID=NEXT\_ID+3
\end{quote}

\textbf{Operation 2}

\textbf{Prompt}
\begin{quote}
\ttfamily
INSTR CALL n=2 src=1 dst=2 aux=3 prev=prev
\end{quote}

\textbf{Target}
\begin{quote}
\ttfamily
APPEND[NEXT\_ID] CALL n=1 src=3 dst=2 aux=1 prev=prev\\
APPEND[NEXT\_ID+1] MOVE d=2 src=1 dst=2 prev=NEXT\_ID\\
APPEND[NEXT\_ID+2] CALL n=1 src=1 dst=3 aux=2 prev=NEXT\_ID+1\\
PTR=NEXT\_ID+2 NEXT\_ID=NEXT\_ID+3
\end{quote}

\textbf{Operation 3}

\textbf{Prompt}
\begin{quote}
\ttfamily
INSTR CALL n=1 src=1 dst=3 aux=2 prev=prev
\end{quote}

\textbf{Target}
\begin{quote}
\ttfamily
OUTPUT Move disk 1 from 1 to 3 PTR=prev NEXT\_ID=NEXT\_ID
\end{quote}

\textbf{Operation 4}

\textbf{Prompt}
\begin{quote}
\ttfamily
INSTR MOVE d=2 src=1 dst=2 prev=prev
\end{quote}

\textbf{Target}
\begin{quote}
\ttfamily
OUTPUT Move disk 2 from 1 to 2 PTR=prev NEXT\_ID=NEXT\_ID
\end{quote}

\textbf{Operation 5}

\textbf{Prompt}
\begin{quote}
\ttfamily
INSTR CALL n=1 src=3 dst=2 aux=1 prev=prev
\end{quote}

\textbf{Target}
\begin{quote}
\ttfamily
OUTPUT Move disk 1 from 3 to 2 PTR=prev NEXT\_ID=NEXT\_ID
\end{quote}

\textbf{Operation 6}

\textbf{Prompt}
\begin{quote}
\ttfamily
INSTR MOVE d=3 src=1 dst=3 prev=prev
\end{quote}

\textbf{Target}
\begin{quote}
\ttfamily
OUTPUT Move disk 3 from 1 to 3 PTR=prev NEXT\_ID=NEXT\_ID
\end{quote}

\textbf{Operation 7}

\textbf{Prompt}
\begin{quote}
\ttfamily
INSTR CALL n=2 src=2 dst=3 aux=1 prev=prev
\end{quote}

\textbf{Target}
\begin{quote}
\ttfamily
APPEND[NEXT\_ID] CALL n=1 src=1 dst=3 aux=2 prev=prev\\
APPEND[NEXT\_ID+1] MOVE d=2 src=2 dst=3 prev=NEXT\_ID\\
APPEND[NEXT\_ID+2] CALL n=1 src=2 dst=1 aux=3 prev=NEXT\_ID+1\\
PTR=NEXT\_ID+2 NEXT\_ID=NEXT\_ID+3
\end{quote}

\textbf{Operation 8}

\textbf{Prompt}
\begin{quote}
\ttfamily
INSTR CALL n=1 src=2 dst=1 aux=3 prev=prev
\end{quote}

\textbf{Target}
\begin{quote}
\ttfamily
OUTPUT Move disk 1 from 2 to 1 PTR=prev NEXT\_ID=NEXT\_ID
\end{quote}

\textbf{Operation 9}

\textbf{Prompt}
\begin{quote}
\ttfamily
INSTR MOVE d=2 src=2 dst=3 prev=prev
\end{quote}

\textbf{Target}
\begin{quote}
\ttfamily
OUTPUT Move disk 2 from 2 to 3 PTR=prev NEXT\_ID=NEXT\_ID
\end{quote}

\end{tcolorbox}
\end{center}

\section{Tower of Hanoi: Complete Example} \label{sec: toh-prompt-example}

In this section, we provide a complete execution of the planning procedure for solving $\textsc{Hanoi}(3,s,d,a)$, including all prompts and the corresponding responses expected from the LM. The \emph{step} variable indicates the order of execution from initialization to termination. Each prompt and target also includes \texttt{NEXT\_ID}, which specifies the current number of instructions in the stack and serves as the index for inserting newly generated instructions during decomposition. 

\vspace{10pt}

\begin{center}
\begin{tcolorbox}[
  colback=gray!12,
  colframe=gray!60,
  colbacktitle=gray!90,
  coltitle=gray!15,
  title={Example: Full Prompt--Target Trace for Tower of Hanoi with $n=3$},
  fonttitle=\bfseries,
  sharp corners,
  boxrule=0.6pt,
  breakable
]
\small

\textbf{Step 1}

\textbf{Prompt}
\begin{quote}
\ttfamily
PTR=0 NEXT\_ID=1 INSTR[0] CALL n=3 src=1 dst=3 aux=2 prev=-1
\end{quote}

\textbf{Target}
\begin{quote}
\ttfamily
APPEND[1] CALL n=2 src=2 dst=3 aux=1 prev=-1\\
APPEND[2] MOVE d=3 src=1 dst=3 prev=1\\
APPEND[3] CALL n=2 src=1 dst=2 aux=3 prev=2\\
PTR=3 NEXT\_ID=4
\end{quote}

\textbf{Step 2}

\textbf{Prompt}
\begin{quote}
\ttfamily
PTR=3 NEXT\_ID=4 INSTR[3] CALL n=2 src=1 dst=2 aux=3 prev=2
\end{quote}

\textbf{Target}
\begin{quote}
\ttfamily
APPEND[4] CALL n=1 src=3 dst=2 aux=1 prev=2\\
APPEND[5] MOVE d=2 src=1 dst=2 prev=4\\
APPEND[6] CALL n=1 src=1 dst=3 aux=2 prev=5\\
PTR=6 NEXT\_ID=7
\end{quote}

\textbf{Step 3}

\textbf{Prompt}
\begin{quote}
\ttfamily
PTR=6 NEXT\_ID=7 INSTR[6] CALL n=1 src=1 dst=3 aux=2 prev=5
\end{quote}

\textbf{Target}
\begin{quote}
\ttfamily
OUTPUT Move disk 1 from 1 to 3\\
PTR=5 NEXT\_ID=7
\end{quote}

\textbf{Step 4}

\textbf{Prompt}
\begin{quote}
\ttfamily
PTR=5 NEXT\_ID=7 INSTR[5] MOVE d=2 src=1 dst=2 prev=4
\end{quote}

\textbf{Target}
\begin{quote}
\ttfamily
OUTPUT Move disk 2 from 1 to 2\\
PTR=4 NEXT\_ID=7
\end{quote}

\textbf{Step 5}

\textbf{Prompt}
\begin{quote}
\ttfamily
PTR=4 NEXT\_ID=7 INSTR[4] CALL n=1 src=3 dst=2 aux=1 prev=2
\end{quote}

\textbf{Target}
\begin{quote}
\ttfamily
OUTPUT Move disk 1 from 3 to 2\\
PTR=2 NEXT\_ID=7
\end{quote}

\textbf{Step 6}

\textbf{Prompt}
\begin{quote}
\ttfamily
PTR=2 NEXT\_ID=7 INSTR[2] MOVE d=3 src=1 dst=3 prev=1
\end{quote}

\textbf{Target}
\begin{quote}
\ttfamily
OUTPUT Move disk 3 from 1 to 3\\
PTR=1 NEXT\_ID=7
\end{quote}

\textbf{Step 7}

\textbf{Prompt}
\begin{quote}
\ttfamily
PTR=1 NEXT\_ID=7 INSTR[1] CALL n=2 src=2 dst=3 aux=1 prev=-1
\end{quote}

\textbf{Target}
\begin{quote}
\ttfamily
APPEND[7] CALL n=1 src=1 dst=3 aux=2 prev=-1\\
APPEND[8] MOVE d=2 src=2 dst=3 prev=7\\
APPEND[9] CALL n=1 src=2 dst=1 aux=3 prev=8\\
PTR=9 NEXT\_ID=10
\end{quote}

\textbf{Step 8}

\textbf{Prompt}
\begin{quote}
\ttfamily
PTR=9 NEXT\_ID=10 INSTR[9] CALL n=1 src=2 dst=1 aux=3 prev=8
\end{quote}

\textbf{Target}
\begin{quote}
\ttfamily
OUTPUT Move disk 1 from 2 to 1\\
PTR=8 NEXT\_ID=10
\end{quote}

\textbf{Step 9}

\textbf{Prompt}
\begin{quote}
\ttfamily
PTR=8 NEXT\_ID=10 INSTR[8] MOVE d=2 src=2 dst=3 prev=7
\end{quote}

\textbf{Target}
\begin{quote}
\ttfamily
OUTPUT Move disk 2 from 2 to 3\\
PTR=7 NEXT\_ID=10
\end{quote}

\textbf{Step 10}

\textbf{Prompt}
\begin{quote}
\ttfamily
PTR=7 NEXT\_ID=10 INSTR[7] CALL n=1 src=1 dst=3 aux=2 prev=-1
\end{quote}

\textbf{Target}
\begin{quote}
\ttfamily
OUTPUT Move disk 1 from 1 to 3\\
PTR=-1 NEXT\_ID=10
\end{quote}

\end{tcolorbox}
\end{center}

\section{Blocksworld: Complete Example} \label{sec: bw-prompt-example}

In this section, we provide a complete example of all prompts and their corresponding targets required to solve a Blocksworld problem with $n=6$ blocks. The initial and goal configurations are sampled uniformly at random from all possible permutations. The \emph{step} variable indicates the order of prompt--target pairs from initialization until the problem is solved. The language model only observes the prompt and generates the target for the next iteration. Whenever the start and goal configurations are identical, the pointer becomes $-1$, indicating that no instructions remain and that the planning algorithm has reached the goal state.

\begin{center}
\begin{tcolorbox}[
  colback=gray!12,
  colframe=gray!60,
  colbacktitle=gray!90,
  coltitle=gray!15,
  title={Example: Full Prompt--Target Trace for Blocksworld with $n=6$},
  fonttitle=\bfseries,
  sharp corners,
  boxrule=0.6pt,
  breakable
]
\small

Initial stack:
\[
[B1, B6, B2, B5, B4, B3]
\]
Goal stack:
\[
[B5, B3, B6, B1, B4, B2]
\]

\textbf{Step 1}

\textbf{Prompt}
\begin{quote}
\ttfamily
PTR=0 CALL type=unstacking start=[B1, B6, B2, B5, B4, B3] \& goal=[B5, B3, B6, B1, B4, B2]
\end{quote}

\textbf{Target}
\begin{quote}
\ttfamily
OUTPUT Move block B3 APPEND[1] CALL type=unstacking start=[B1, B6, B2, B5, B4] \& goal=[B5, B3, B6, B1, B4, B2] PTR=1
\end{quote}

\textbf{Step 2}

\textbf{Prompt}
\begin{quote}
\ttfamily
PTR=1 CALL type=unstacking start=[B1, B6, B2, B5, B4] \& goal=[B5, B3, B6, B1, B4, B2]
\end{quote}

\textbf{Target}
\begin{quote}
\ttfamily
OUTPUT Move block B4 APPEND[2] CALL type=unstacking start=[B1, B6, B2, B5] \& goal=[B5, B3, B6, B1, B4, B2] PTR=2
\end{quote}

\textbf{Step 3}

\textbf{Prompt}
\begin{quote}
\ttfamily
PTR=2 CALL type=unstacking start=[B1, B6, B2, B5] \& goal=[B5, B3, B6, B1, B4, B2]
\end{quote}

\textbf{Target}
\begin{quote}
\ttfamily
OUTPUT Move block B5 APPEND[3] CALL type=unstacking start=[B1, B6, B2] \& goal=[B5, B3, B6, B1, B4, B2] PTR=3
\end{quote}

\textbf{Step 4}

\textbf{Prompt}
\begin{quote}
\ttfamily
PTR=3 CALL type=unstacking start=[B1, B6, B2] \& goal=[B5, B3, B6, B1, B4, B2]
\end{quote}

\textbf{Target}
\begin{quote}
\ttfamily
OUTPUT Move block B2 APPEND[4] CALL type=unstacking start=[B1, B6] \& goal=[B5, B3, B6, B1, B4, B2] PTR=4
\end{quote}

\textbf{Step 5}

\textbf{Prompt}
\begin{quote}
\ttfamily
PTR=4 CALL type=unstacking start=[B1, B6] \& goal=[B5, B3, B6, B1, B4, B2]
\end{quote}

\textbf{Target}
\begin{quote}
\ttfamily
OUTPUT Move block B6 APPEND[5] CALL type=unstacking start=[B1] \& goal=[B5, B3, B6, B1, B4, B2] PTR=5
\end{quote}

\textbf{Step 6}

\textbf{Prompt}
\begin{quote}
\ttfamily
PTR=5 CALL type=unstacking start=[B1] \& goal=[B5, B3, B6, B1, B4, B2]
\end{quote}

\textbf{Target}
\begin{quote}
\ttfamily
OUTPUT Move block B1 APPEND[6] CALL type=stacking start=[] \& goal=[B5, B3, B6, B1, B4, B2] PTR=6
\end{quote}

\textbf{Step 7}

\textbf{Prompt}
\begin{quote}
\ttfamily
PTR=6 CALL type=stacking start=[] \& goal=[B5, B3, B6, B1, B4, B2]
\end{quote}

\textbf{Target}
\begin{quote}
\ttfamily
OUTPUT Move block B5 APPEND[7] CALL type=stacking start=[B5] \& goal=[B5, B3, B6, B1, B4, B2] PTR=7
\end{quote}

\textbf{Step 8}

\textbf{Prompt}
\begin{quote}
\ttfamily
PTR=7 CALL type=stacking start=[B5] \& goal=[B5, B3, B6, B1, B4, B2]
\end{quote}

\textbf{Target}
\begin{quote}
\ttfamily
OUTPUT Move block B3 APPEND[8] CALL type=stacking start=[B5, B3] \& goal=[B5, B3, B6, B1, B4, B2] PTR=8
\end{quote}

\textbf{Step 9}

\textbf{Prompt}
\begin{quote}
\ttfamily
PTR=8 CALL type=stacking start=[B5, B3] \& goal=[B5, B3, B6, B1, B4, B2]
\end{quote}

\textbf{Target}
\begin{quote}
\ttfamily
OUTPUT Move block B6 APPEND[9] CALL type=stacking start=[B5, B3, B6] \& goal=[B5, B3, B6, B1, B4, B2] PTR=9
\end{quote}

\textbf{Step 10}

\textbf{Prompt}
\begin{quote}
\ttfamily
PTR=9 CALL type=stacking start=[B5, B3, B6] \& goal=[B5, B3, B6, B1, B4, B2]
\end{quote}

\textbf{Target}
\begin{quote}
\ttfamily
OUTPUT Move block B1 APPEND[10] CALL type=stacking start=[B5, B3, B6, B1] \& goal=[B5, B3, B6, B1, B4, B2] PTR=10
\end{quote}

\textbf{Step 11}

\textbf{Prompt}
\begin{quote}
\ttfamily
PTR=10 CALL type=stacking start=[B5, B3, B6, B1] \& goal=[B5, B3, B6, B1, B4, B2]
\end{quote}

\textbf{Target}
\begin{quote}
\ttfamily
OUTPUT Move block B4 APPEND[11] CALL type=stacking start=[B5, B3, B6, B1, B4] \& goal=[B5, B3, B6, B1, B4, B2] PTR=11
\end{quote}

\textbf{Step 12}

\textbf{Prompt}
\begin{quote}
\ttfamily
PTR=11 CALL type=stacking start=[B5, B3, B6, B1, B4] \& goal=[B5, B3, B6, B1, B4, B2]
\end{quote}

\textbf{Target}
\begin{quote}
\ttfamily
OUTPUT Move block B2 APPEND[-1] CALL type=stacking start=[B5, B3, B6, B1, B4, B2] \& goal=[B5, B3, B6, B1, B4, B2] PTR=-1
\end{quote}

\end{tcolorbox}
\end{center}

\section{Pancake Puzzle: Complete Example} \label{sec: pancake-prompt-example}

In this section, we provide a complete example of all prompts and their corresponding targets required to solve a Pancake Puzzle problem with $n=6$ pancakes. The initial configuration is sampled uniformly at random from all possible permutations, while the goal configuration is the same sorted order for all problem instances. The \emph{step} variable indicates the order of prompt--target pairs from initialization until the problem is solved. At each step, two flip moves are generated to place the next largest unsorted pancake into its correct position. The pointer to the next instruction for the following iteration is generated in the target and always refers to the next instruction in the context window. Whenever the start and goal configurations are identical, the pointer becomes $-1$, indicating that no instructions remain and that the planning algorithm has reached the goal state.

\begin{center}
\begin{tcolorbox}[
  colback=gray!12,
  colframe=gray!60,
  colbacktitle=gray!90,
  coltitle=gray!15,
  title={Example: Full Prompt--Target Trace for Pancake Puzzle with $n=6$},
  fonttitle=\bfseries,
  sharp corners,
  boxrule=0.6pt,
  breakable
]
\small

Initial state:
\[
[P4, P6, P2, P5, P1, P3]
\]
Goal state:
\[
[P1, P2, P3, P4, P5, P6]
\]

\textbf{Step 1}

\textbf{Prompt}
\begin{quote}
\ttfamily
CALL instruction pointer=0 state=[P4, P6, P2, P5, P1, P3] \& goal=[P1, P2, P3, P4, P5, P6]
\end{quote}

\textbf{Target}
\begin{quote}
\ttfamily
The largest unsorted pancake is P1.\\
OUTPUT Flip from where P1 is located.\\
State after first flip: [P4, P6, P2, P5, P3, P1].\\
OUTPUT Flip from where P1 must be located.\\
State after second flip: [P1, P3, P5, P2, P6, P4].\\
APPEND[1] CALL instruction state=[P1, P3, P5, P2, P6, P4] \& goal=[P1, P2, P3, P4, P5, P6] pointer=1.
\end{quote}

\textbf{Step 2}

\textbf{Prompt}
\begin{quote}
\ttfamily
CALL instruction pointer=1 state=[P1, P3, P5, P2, P6, P4] \& goal=[P1, P2, P3, P4, P5, P6]
\end{quote}

\textbf{Target}
\begin{quote}
\ttfamily
The largest unsorted pancake is P2.\\
OUTPUT Flip from where P2 is located.\\
State after first flip: [P1, P3, P5, P4, P6, P2].\\
OUTPUT Flip from where P2 must be located.\\
State after second flip: [P1, P2, P6, P4, P5, P3].\\
APPEND[2] CALL instruction state=[P1, P2, P6, P4, P5, P3] \& goal=[P1, P2, P3, P4, P5, P6] pointer=2.
\end{quote}

\textbf{Step 3}

\textbf{Prompt}
\begin{quote}
\ttfamily
CALL instruction pointer=2 state=[P1, P2, P6, P4, P5, P3] \& goal=[P1, P2, P3, P4, P5, P6]
\end{quote}

\textbf{Target}
\begin{quote}
\ttfamily
The largest unsorted pancake is P3.\\
OUTPUT Flip from where P3 is located.\\
State after first flip: [P1, P2, P6, P4, P5, P3].\\
OUTPUT Flip from where P3 must be located.\\
State after second flip: [P1, P2, P3, P5, P4, P6].\\
APPEND[3] CALL instruction state=[P1, P2, P3, P5, P4, P6] \& goal=[P1, P2, P3, P4, P5, P6] pointer=3.
\end{quote}

\textbf{Step 4}

\textbf{Prompt}
\begin{quote}
\ttfamily
CALL instruction pointer=3 state=[P1, P2, P3, P5, P4, P6] \& goal=[P1, P2, P3, P4, P5, P6]
\end{quote}

\textbf{Target}
\begin{quote}
\ttfamily
The largest unsorted pancake is P4.\\
OUTPUT Flip from where P4 is located.\\
State after first flip: [P1, P2, P3, P5, P6, P4].\\
OUTPUT Flip from where P4 must be located.\\
State after second flip: [P1, P2, P3, P4, P6, P5].\\
APPEND[4] CALL instruction state=[P1, P2, P3, P4, P6, P5] \& goal=[P1, P2, P3, P4, P5, P6] pointer=4.
\end{quote}

\textbf{Step 5}

\textbf{Prompt}
\begin{quote}
\ttfamily
CALL instruction pointer=4 state=[P1, P2, P3, P4, P6, P5] \& goal=[P1, P2, P3, P4, P5, P6]
\end{quote}

\textbf{Target}
\begin{quote}
\ttfamily
The largest unsorted pancake is P5.\\
OUTPUT Flip from where P5 is located.\\
State after first flip: [P1, P2, P3, P4, P6, P5].\\
OUTPUT Flip from where P5 must be located.\\
State after second flip: [P1, P2, P3, P4, P5, P6].\\
APPEND[-1] CALL instruction state=[P1, P2, P3, P4, P5, P6] \& goal=[P1, P2, P3, P4, P5, P6] pointer=-1.
\end{quote}

\end{tcolorbox}
\end{center}

\section{Extended Blocksworld: Complete Example} \label{sec: extended-blocksworld-prompt-example}

In this section, we provide a complete example of all prompts and their corresponding targets required to solve an \emph{Extended BlocksWorld} problem with $n=6$ blocks. The initial and goal configurations are sampled uniformly at random from all possible permutations. In this setting, the number of stacks is also sampled from $1$ to $n$, with the constraint that no stack is empty. This is because blocks can be placed directly on the table, and an empty stack is equivalent to the table. The \emph{step} variable indicates the order of prompt--target pairs from initialization until the problem is solved. The language model observes only the prompt and generates the target for the next iteration. Whenever the start and goal configurations are identical, the pointer becomes $-1$, indicating that no instructions remain and that the planning algorithm has reached the goal state. The language model always attempts to match the goal configuration in the first stack of the start state and leaves the remaining stacks empty.

\begin{center}
\begin{tcolorbox}[
  colback=gray!12,
  colframe=gray!60,
  colbacktitle=gray!90,
  coltitle=gray!15,
  title={Example: Full Prompt--Target Trace for Extended BlocksWorld with $n=6$},
  fonttitle=\bfseries,
  sharp corners,
  boxrule=0.6pt,
  breakable
]
\small

Initial stacks:
\[
[[B1, B6], [B2, B5, B4], [B3]]
\]
Goal stack:
\[
[B5, B3, B6, B1, B4, B2]
\]

\textbf{Step 1}

\textbf{Prompt}
\begin{quote}
\ttfamily
PTR=0 CALL type=unstacking start=[[B1, B6], [B2, B5, B4], [B3]] \& goal=[B5, B3, B6, B1, B4, B2]
\end{quote}

\textbf{Target}
\begin{quote}
\ttfamily
OUTPUT Move block B3 APPEND[1] CALL type=unstacking start=[[B1, B6], [B2, B5, B4]] \& goal=[B5, B3, B6, B1, B4, B2] PTR=1
\end{quote}

\textbf{Step 2}

\textbf{Prompt}
\begin{quote}
\ttfamily
PTR=1 CALL type=unstacking start=[[B1, B6], [B2, B5, B4]] \& goal=[B5, B3, B6, B1, B4, B2]
\end{quote}

\textbf{Target}
\begin{quote}
\ttfamily
OUTPUT Move block B4 APPEND[2] CALL type=unstacking start=[[B1, B6], [B2, B5]] \& goal=[B5, B3, B6, B1, B4, B2] PTR=2
\end{quote}

\textbf{Step 3}

\textbf{Prompt}
\begin{quote}
\ttfamily
PTR=2 CALL type=unstacking start=[[B1, B6], [B2, B5]] \& goal=[B5, B3, B6, B1, B4, B2]
\end{quote}

\textbf{Target}
\begin{quote}
\ttfamily
OUTPUT Move block B5 APPEND[3] CALL type=unstacking start=[[B1, B6], [B2]] \& goal=[B5, B3, B6, B1, B4, B2] PTR=3
\end{quote}

\textbf{Step 4}

\textbf{Prompt}
\begin{quote}
\ttfamily
PTR=3 CALL type=unstacking start=[[B1, B6], [B2]] \& goal=[B5, B3, B6, B1, B4, B2]
\end{quote}

\textbf{Target}
\begin{quote}
\ttfamily
OUTPUT Move block B2 APPEND[4] CALL type=unstacking start=[[B1, B6]] \& goal=[B5, B3, B6, B1, B4, B2] PTR=4
\end{quote}

\textbf{Step 5}

\textbf{Prompt}
\begin{quote}
\ttfamily
PTR=4 CALL type=unstacking start=[[B1, B6]] \& goal=[B5, B3, B6, B1, B4, B2]
\end{quote}

\textbf{Target}
\begin{quote}
\ttfamily
OUTPUT Move block B6 APPEND[5] CALL type=unstacking start=[[B1]] \& goal=[B5, B3, B6, B1, B4, B2] PTR=5
\end{quote}

\textbf{Step 6}

\textbf{Prompt}
\begin{quote}
\ttfamily
PTR=5 CALL type=unstacking start=[[B1]] \& goal=[B5, B3, B6, B1, B4, B2]
\end{quote}

\textbf{Target}
\begin{quote}
\ttfamily
OUTPUT Move block B1 APPEND[6] CALL type=stacking start=[] \& goal=[B5, B3, B6, B1, B4, B2] PTR=6
\end{quote}

\textbf{Step 7}

\textbf{Prompt}
\begin{quote}
\ttfamily
PTR=6 CALL type=stacking start=[] \& goal=[B5, B3, B6, B1, B4, B2]
\end{quote}

\textbf{Target}
\begin{quote}
\ttfamily
OUTPUT Move block B5 APPEND[7] CALL type=stacking start=[[B5]] \& goal=[B5, B3, B6, B1, B4, B2] PTR=7
\end{quote}

\textbf{Step 8}

\textbf{Prompt}
\begin{quote}
\ttfamily
PTR=7 CALL type=stacking start=[[B5]] \& goal=[B5, B3, B6, B1, B4, B2]
\end{quote}

\textbf{Target}
\begin{quote}
\ttfamily
OUTPUT Move block B3 APPEND[8] CALL type=stacking start=[[B5, B3]] \& goal=[B5, B3, B6, B1, B4, B2] PTR=8
\end{quote}

\textbf{Step 9}

\textbf{Prompt}
\begin{quote}
\ttfamily
PTR=8 CALL type=stacking start=[[B5, B3]] \& goal=[B5, B3, B6, B1, B4, B2]
\end{quote}

\textbf{Target}
\begin{quote}
\ttfamily
OUTPUT Move block B6 APPEND[9] CALL type=stacking start=[[B5, B3, B6]] \& goal=[B5, B3, B6, B1, B4, B2] PTR=9
\end{quote}

\textbf{Step 10}

\textbf{Prompt}
\begin{quote}
\ttfamily
PTR=9 CALL type=stacking start=[[B5, B3, B6]] \& goal=[B5, B3, B6, B1, B4, B2]
\end{quote}

\textbf{Target}
\begin{quote}
\ttfamily
OUTPUT Move block B1 APPEND[10] CALL type=stacking start=[[B5, B3, B6, B1]] \& goal=[B5, B3, B6, B1, B4, B2] PTR=10
\end{quote}

\textbf{Step 11}

\textbf{Prompt}
\begin{quote}
\ttfamily
PTR=10 CALL type=stacking start=[[B5, B3, B6, B1]] \& goal=[B5, B3, B6, B1, B4, B2]
\end{quote}

\textbf{Target}
\begin{quote}
\ttfamily
OUTPUT Move block B4 APPEND[11] CALL type=stacking start=[[B5, B3, B6, B1, B4]] \& goal=[B5, B3, B6, B1, B4, B2] PTR=11
\end{quote}

\textbf{Step 12}

\textbf{Prompt}
\begin{quote}
\ttfamily
PTR=11 CALL type=stacking start=[[B5, B3, B6, B1, B4]] \& goal=[B5, B3, B6, B1, B4, B2]
\end{quote}

\textbf{Target}
\begin{quote}
\ttfamily
OUTPUT Move block B2 APPEND[-1] CALL type=stacking start=[[B5, B3, B6, B1, B4, B2]] \& goal=[B5, B3, B6, B1, B4, B2] PTR=-1
\end{quote}

\end{tcolorbox}
\end{center}

\section{PDA for Tower of Hanoi: Complete Example} \label{sec: TOH as PDA}

In this section, we first show how the planner for TOH can resemble a PDA. A PDA is formally defined as a tuple
\[
\mathcal{A}
=
\left(
Q,\Sigma,\Gamma,\delta,q_0,Z_0,F
\right),
\]
where:
\begin{itemize}
\item $Q$ is a finite set of control states;

\item $\Sigma$ is the input alphabet, containing the symbols that may
appear in the input sequence;

\item $\Gamma$ is the stack alphabet, containing the symbols that may be
stored in the pushdown stack;

\item $\delta$ is the transition function, which determines the possible
next control states and stack updates;

\item $q_0 \in Q$ is the initial control state;

\item $Z_0 \in \Gamma$ is the initial stack symbol, typically used as the
bottom-of-stack marker;

\item $F \subseteq Q$ is the set of accepting states.

\end{itemize}

Assume that the number of disks is bounded by a fixed constant \(N\), and let \(P=\{1,2,3\}\) denote the set of pegs. We represent a recursive instruction by $C(k,s,d,a)$, meaning that \(k\) disks must be moved from source peg \(s\) to destination peg \(d\) using auxiliary peg \(a\), and an explicit move instruction by $M(k,s,d),$ meaning that disk \(k\) must be moved from peg \(s\) to peg \(d\). The planner can be represented by the deterministic pushdown automaton where
\[
Q=\{q_{\mathrm{init}},q_{\mathrm{run}},q_{\mathrm{acc}}\},
\qquad
F=\{q_{\mathrm{acc}}\},
\]
and
\[
\Sigma_N=\{\tau_n \mid 1\leq n\leq N\},
\]
with \(\tau_n\) denoting a Tower-of-Hanoi instance containing \(n\) disks.

The stack alphabet is
\[
\Gamma_N
=
\{Z_0\}
\cup
\{C(k,s,d,a)\}
\cup
\{M(k,s,d)\},
\]
where \(1\leq k\leq N\), \(s,d,a\in P\), the pegs in a call instruction are
pairwise distinct, and \(s\neq d\) in a move instruction. Since \(N\) is
bounded, \(\Gamma_N\) is finite.

The transition function is defined as follows.

\emph{Initialization:}
\[
\delta(q_{\mathrm{init}},\tau_n,Z_0)
=
\bigl(q_{\mathrm{run}},C(n,1,3,2)Z_0\bigr).
\]

\emph{Recursive expansion, for \(k>1\):}
\[
\delta(q_{\mathrm{run}},\epsilon,C(k,s,d,a))
=
\Bigl(
q_{\mathrm{run}},
C(k-1,s,a,d)\,
M(k,s,d)\,
C(k-1,a,d,s)
\Bigr).
\]

\emph{Base case:}
\[
\delta(q_{\mathrm{run}},\epsilon,C(1,s,d,a))
=
(q_{\mathrm{run}},\epsilon).
\]

\emph{Move execution:}
\[
\delta(q_{\mathrm{run}},\epsilon,M(k,s,d))
=
(q_{\mathrm{run}},\epsilon).
\]

\emph{Termination:}
\[
\delta(q_{\mathrm{run}},\epsilon,Z_0)
=
(q_{\mathrm{acc}},Z_0).
\]

Thus, recursive calls are replaced by three unfinished instructions, whereas
base-case calls and explicit move instructions are popped from the stack.
The planner accepts when all instructions have been processed and only the
bottom-of-stack marker remains.

Because the planner also emits actions, its complete formulation is more
precisely a deterministic pushdown transducer \citep{aho1969syntax}. Let
\[
\Omega_N
=
\{\operatorname{Move}(k,s,d)
\mid
1\leq k\leq N,\ s,d\in P,\ s\neq d\}
\]
be the output alphabet. The output function satisfies
\[
\lambda(q_{\mathrm{run}},\epsilon,C(1,s,d,a))
=
\operatorname{Move}(1,s,d),
\]
\[
\lambda(q_{\mathrm{run}},\epsilon,M(k,s,d))
=
\operatorname{Move}(k,s,d),
\]
and produces no output for recursive-call expansion.

Below, we show the complete sequence of prompt--target pairs for the planning procedure with $n=3$ disks. The pop operation is omitted for clarity. Each time a prompt is presented to the LM, it is implicitly assumed that the corresponding instruction has been popped from the top of the stack. The output transduction performed by the $\lambda$ function is indicated by the \texttt{OUTPUT} token preceding each action.

\vspace{10pt}

\begin{center}
\begin{tcolorbox}[
  colback=gray!12,
  colframe=gray!60,
  colbacktitle=gray!90,
  coltitle=gray!15,
  title={Example: Full Push--Pop Prompt--Target Trace for Tower of Hanoi with $n=3$},
  fonttitle=\bfseries,
  sharp corners,
  boxrule=0.6pt,
  breakable
]
\small

\textbf{Step 1}

\textbf{Prompt}
\begin{quote}
\ttfamily
INSTR CALL n=3 src=1 dst=3 aux=2
\end{quote}

\textbf{Target}
\begin{quote}
\ttfamily
PUSH CALL n=2 src=2 dst=3 aux=1\\
PUSH MOVE d=3 src=1 dst=3\\
PUSH CALL n=2 src=1 dst=2 aux=3
\end{quote}

\textbf{Step 2}

\textbf{Prompt}
\begin{quote}
\ttfamily
INSTR CALL n=2 src=1 dst=2 aux=3
\end{quote}

\textbf{Target}
\begin{quote}
\ttfamily
PUSH CALL n=1 src=3 dst=2 aux=1\\
PUSH MOVE d=2 src=1 dst=2\\
PUSH CALL n=1 src=1 dst=3 aux=2
\end{quote}

\textbf{Step 3}

\textbf{Prompt}
\begin{quote}
\ttfamily
INSTR CALL n=1 src=1 dst=3 aux=2
\end{quote}

\textbf{Target}
\begin{quote}
\ttfamily
OUTPUT Move disk 1 from 1 to 3
\end{quote}

\textbf{Step 4}

\textbf{Prompt}
\begin{quote}
\ttfamily
INSTR MOVE d=2 src=1 dst=2
\end{quote}

\textbf{Target}
\begin{quote}
\ttfamily
OUTPUT Move disk 2 from 1 to 2
\end{quote}

\textbf{Step 5}

\textbf{Prompt}
\begin{quote}
\ttfamily
INSTR CALL n=1 src=3 dst=2 aux=1
\end{quote}

\textbf{Target}
\begin{quote}
\ttfamily
OUTPUT Move disk 1 from 3 to 2
\end{quote}

\textbf{Step 6}

\textbf{Prompt}
\begin{quote}
\ttfamily
INSTR MOVE d=3 src=1 dst=3
\end{quote}

\textbf{Target}
\begin{quote}
\ttfamily
OUTPUT Move disk 3 from 1 to 3
\end{quote}

\textbf{Step 7}

\textbf{Prompt}
\begin{quote}
\ttfamily
INSTR CALL n=2 src=2 dst=3 aux=1
\end{quote}

\textbf{Target}
\begin{quote}
\ttfamily
PUSH CALL n=1 src=1 dst=3 aux=2\\
PUSH MOVE d=2 src=2 dst=3\\
PUSH CALL n=1 src=2 dst=1 aux=3
\end{quote}

\textbf{Step 8}

\textbf{Prompt}
\begin{quote}
\ttfamily
INSTR CALL n=1 src=2 dst=1 aux=3
\end{quote}

\textbf{Target}
\begin{quote}
\ttfamily
OUTPUT Move disk 1 from 2 to 1
\end{quote}

\textbf{Step 9}

\textbf{Prompt}
\begin{quote}
\ttfamily
INSTR MOVE d=2 src=2 dst=3
\end{quote}

\textbf{Target}
\begin{quote}
\ttfamily
OUTPUT Move disk 2 from 2 to 3
\end{quote}

\textbf{Step 10}

\textbf{Prompt}
\begin{quote}
\ttfamily
INSTR CALL n=1 src=1 dst=3 aux=2
\end{quote}

\textbf{Target}
\begin{quote}
\ttfamily
OUTPUT Move disk 1 from 1 to 3
\end{quote}

\end{tcolorbox}
\end{center}

\section{PDA formulation for Depth-first search} \label{sec: pda-dfs}

In this section, we show that the depth-first search (DFS) algorithm can be formulated as a deterministic PDA. We assume that the complexity of every planning instance can be measured and is bounded by $N$. Then, we can define the family of planning instances as follows:
\[
\Sigma_N
=
\left\{
\tau
\mid
\tau \text{ is a planning instance of complexity at most } N
\right\}.
\]
Each planning instance \(\tau\in\Sigma_N\) specifies an initial state \(s_0^\tau\), a set of goal states \(G_\tau\), and a world model
\(\mathcal{T}_\tau(s'\mid s,a)\) that generates the successor states obtained by applying an action \(a\) in state \(s\). We consider the tree-search version of DFS and assume that the search depth and branching factor are bounded. Let \(\mathcal{V}_\tau\) denote the finite set of
search nodes that may be generated for an instance \(\tau\). Each node \(v\in\mathcal{V}_\tau\) contains a world state \(s(v)\). For each node \(v\),
let
\[
\operatorname{Succ}_\tau(v)
=
\langle v_1,\ldots,v_m\rangle
\]
denote its ordered sequence of child nodes, generated using the world model
\(\mathcal{T}_\tau\). We assume a fixed ordering of the actions and, consequently, of the generated child nodes. The DFS algorithm can be represented by the PDA
\[
Q
=
\{
q_{\mathrm{init}},
q_{\mathrm{run}},
q_{\mathrm{acc}},
q_{\mathrm{rej}}
\},
\qquad
F=\{q_{\mathrm{acc}}\}.
\]

The stack alphabet is
\[
\Gamma_N
=
\{Z_0\}
\cup
\{
V(\tau,v)
\mid
\tau\in\Sigma_N,\;
v\in\mathcal{V}_\tau
\},
\]
where \(V(\tau,v)\) denotes a search node \(v\) in the search tree corresponding to the planning instance $\tau$. Because the planning instances, branching factor, and search depth are bounded, \(\Gamma_N\) is finite. The transition function is defined as follows.

\emph{Initialization:}
\[
\delta
\bigl(
q_{\mathrm{init}},
\tau,
Z_0
\bigr)
=
\bigl(
q_{\mathrm{run}},
V(\tau,v_0^\tau)Z_0
\bigr),
\]
where \(v_0^\tau\) is the root node satisfying
\[
s(v_0^\tau)=s_0^\tau.
\]

\emph{Goal detection:}
\[
\delta
\bigl(
q_{\mathrm{run}},
\epsilon,
V(\tau,v)
\bigr)
=
\bigl(
q_{\mathrm{acc}},
\epsilon
\bigr),
\qquad
\text{if } s(v)\in G_\tau.
\]

\emph{Node expansion:}
if \(s(v)\notin G_\tau\) and
\[
\operatorname{Succ}_\tau(v)
=
\langle v_1,\ldots,v_m\rangle,
\qquad
m>0,
\]
then
\[
\delta
\bigl(
q_{\mathrm{run}},
\epsilon,
V(\tau,v)
\bigr)
=
\Bigl(
q_{\mathrm{run}},
V(\tau,v_1)
V(\tau,v_2)
\cdots
V(\tau,v_m)
\Bigr).
\]

\emph{Dead-end removal:}
\[
\delta
\bigl(
q_{\mathrm{run}},
\epsilon,
V(\tau,v)
\bigr)
=
\bigl(
q_{\mathrm{run}},
\epsilon
\bigr),
\]
if
\[
s(v)\notin G_\tau
\qquad\text{and}\qquad
\operatorname{Succ}_\tau(v)=\langle\rangle.
\]

\emph{Search failure:}
\[
\delta
\bigl(
q_{\mathrm{run}},
\epsilon,
Z_0
\bigr)
=
\bigl(
q_{\mathrm{rej}},
Z_0
\bigr).
\]

Thus, at each iteration, DFS processes the node at the top of the stack. If the node satisfies the goal condition, the automaton immediately enters the
accepting state and terminates, regardless of whether other unexplored nodes remain on the stack. Otherwise, the node is replaced by its ordered children. Because the first child is placed at the top of the stack, its subtree is explored before the remaining children, reproducing the depth-first search order. If a node has no successors, it is popped from the stack, causing the search to backtrack to the next unexplored node. If the stack is exhausted without reaching a goal state, the automaton enters the rejecting state. Unlike the TOH planner, this DFS formulation does not emit any actions or other outputs. Therefore, no output alphabet or output function is required, and DFS is represented as a deterministic pushdown automaton rather than a pushdown transducer.

\subsection{Training and Evaluation Details}

The model was trained on 120 mazes, with 40 mazes from each of the $11\times11$, $21\times21$, and $31\times31$ size groups and 100 randomly generated start--goal DFS simulations per maze. Testing was conducted on 45 unseen mazes, with 15 mazes of each size and 10 start--goal pairs evaluated per maze. Relative to the model architecture used for the other planning domains, the only change was reducing the hidden size $d_{\mathrm{model}}$ from 512 to 256; all other architecture and optimization settings were retained. Each input token is represented as the sum of a token embedding, a learned one-dimensional absolute position embedding, learned row and column embeddings encoding its maze coordinates, a segment embedding distinguishing the prompt maze from up to four child mazes, and an edit/copy-region embedding. The edit/copy-region embedding marks the current location \texttt{X} and its four cardinal neighbors in the prompt as editable, while all remaining maze cells are marked as copy regions.

\subsection{Prompt--Target Examples for an LM Trained from Scratch} \label{sec: prompt-target for DFS-LM}

Each search state is represented as a complete maze grid using the symbols \texttt{@}, \texttt{.}, \texttt{X}, \texttt{O}, \texttt{G}, and \texttt{\$}, denoting walls, unvisited cells, the current state, states on the current root-to-node path, the goal, and a reached goal, respectively. No coordinates or textual state descriptions are included. Given the grid, the LM generates all valid unvisited children in the fixed order down, right, up, and left. Each child is preceded by the token \texttt{Push} and represented by the complete resulting maze, where the current \texttt{X} becomes \texttt{O}, the selected neighboring cell becomes \texttt{X}, and all existing \texttt{O} markers are retained. If the selected child is the goal, \texttt{X} and \texttt{G} are replaced by \texttt{\$}. If no valid child exists, the model generates the \texttt{EOS} token immediately. Below, we provide two examples: one with multiple children and another with none.

\vspace{10pt}

\begin{center}
\begin{tcolorbox}[
  colback=gray!12,
  colframe=gray!60,
  colbacktitle=gray!90,
  coltitle=gray!15,
  title={Example 1: DFS State with Multiple Valid Children},
  fonttitle=\bfseries,
  sharp corners,
  boxrule=0.6pt,
  breakable
]
\small

\textbf{Prompt}
\begin{quote}
\ttfamily
@@@@@@@@@@@\\
@..G......@\\
@.@@@.@@@@@\\
@.@.@.....@\\
@@@.@@@@@.@\\
@...@.....@\\
@.@.@@@.@.@\\
@.@.@..X@.@\\
@.@@@.@.@.@\\
@.....@.@.@\\
@@@@@@@@@@@
\end{quote}

\vspace{5pt}

\textbf{Target}
\begin{quote}
\ttfamily
Push\\
@@@@@@@@@@@\\
@..G......@\\
@.@@@.@@@@@\\
@.@.@.....@\\
@@@.@@@@@.@\\
@...@.....@\\
@.@.@@@.@.@\\
@.@.@..O@.@\\
@.@@@.@X@.@\\
@.....@.@.@\\
@@@@@@@@@@@\\
Push\\
@@@@@@@@@@@\\
@..G......@\\
@.@@@.@@@@@\\
@.@.@.....@\\
@@@.@@@@@.@\\
@...@.....@\\
@.@.@@@X@.@\\
@.@.@..O@.@\\
@.@@@.@.@.@\\
@.....@.@.@\\
@@@@@@@@@@@\\
Push\\
@@@@@@@@@@@\\
@..G......@\\
@.@@@.@@@@@\\
@.@.@.....@\\
@@@.@@@@@.@\\
@...@.....@\\
@.@.@@@.@.@\\
@.@.@.XO@.@\\
@.@@@.@.@.@\\
@.....@.@.@\\
@@@@@@@@@@@
\end{quote}

\end{tcolorbox}
\end{center}

\vspace{10pt}

\begin{center}
\begin{tcolorbox}[
  colback=gray!12,
  colframe=gray!60,
  colbacktitle=gray!90,
  coltitle=gray!15,
  title={Example 2: DFS State with No Valid Children},
  fonttitle=\bfseries,
  sharp corners,
  boxrule=0.6pt,
  breakable
]
\small

\textbf{Prompt}
\begin{quote}
\ttfamily
@@@@@@@@@@@\\
@..G......@\\
@.@@@.@@@@@\\
@.@X@.....@\\
@@@O@@@@@.@\\
@OOO@.....@\\
@O@.@@@.@.@\\
@O@.@OOO@.@\\
@O@@@O@.@.@\\
@OOOOO@.@.@\\
@@@@@@@@@@@
\end{quote}

\vspace{5pt}

\textbf{Target}
\begin{quote}
\emph{Empty string; EOS is generated immediately.}
\end{quote}

\end{tcolorbox}
\end{center}

\subsection{Prompt--Target Examples for a Pretrained LM} \label{sec: prompt-target for DFS-Qwen}

In this section, we present an example for each prompting mode used with Qwen3.5-9B. The third mode, designed for the full-context setup, is omitted due to its length, as it simply extends the first mode by retaining all states encountered during planning. We also omit the target for each example, since it is straightforward and follows the same structure as in Section~\ref{sec: prompt-target for DFS-LM}.

\begin{tcolorbox}[
  colback=gray!12,
  colframe=gray!60,
  colbacktitle=gray!90,
  coltitle=gray!15,
  title={Qwen: Zero-Shot Prompt},
  fonttitle=\bfseries,
  sharp corners,
  boxrule=0.6pt,
  breakable
]
\small

\textbf{System message}

You perform one child-expansion step for a maze state.
The input is a rectangular character grid:
\texttt{@} = wall,
\texttt{.} = open cell,
\texttt{X} = current cell,
\texttt{G} = goal,
\texttt{O} = cells already visited on the current path,
and \texttt{\$} = the current cell coinciding with the goal.

Generate ALL legal children of \texttt{X} in this exact order:
down, right, up, left.
A legal child is one orthogonal step inside the grid onto an
open cell (\texttt{.}) or the goal (\texttt{G}).
Never move onto a wall (\texttt{@}) or a visited cell (\texttt{O}),
outside the grid, or diagonally.

For EACH legal child, output a line containing exactly
\texttt{Push} followed by the ENTIRE child grid.
To construct each child grid:
\begin{enumerate}
  \item Start with a copy of the input grid.
  \item Mark the current cell (\texttt{X}) as visited (\texttt{O}).
  \item Mark this child's destination as the new current cell
        (\texttt{X}). If the destination is the goal (\texttt{G}),
        mark it with \texttt{\$} instead.
  \item Leave every other cell unchanged, including previously
        visited cells (\texttt{O}) and the goal (\texttt{G})
        when it is not the destination.
\end{enumerate}

If there are no legal children, return an empty response:
generate no text and end your turn.
Return only the \texttt{Push}/grid blocks, or nothing:
no explanation, coordinates, numbering, Markdown fences,
or blank lines between blocks.

\medskip
\textbf{User message}

State:
\begin{quote}
\ttfamily
@@@@@@@@@@@\\
@..G......@\\
@.@@@.@@@@@\\
@.@.@.....@\\
@@@.@@@@@.@\\
@...@.....@\\
@.@.@@@.@.@\\
@.@.@..X@.@\\
@.@@@.@.@.@\\
@.....@.@.@\\
@@@@@@@@@@@
\end{quote}

\end{tcolorbox}

\begin{tcolorbox}[
  colback=gray!12,
  colframe=gray!60,
  colbacktitle=gray!90,
  coltitle=gray!15,
  title={Qwen: One-Shot Prompt},
  fonttitle=\bfseries,
  sharp corners,
  boxrule=0.6pt,
  breakable
]
\small

\textbf{System message}

You perform one child-expansion step for a maze state.
The input is a rectangular character grid:
\texttt{@} = wall,
\texttt{.} = open cell,
\texttt{X} = current cell,
\texttt{G} = goal,
\texttt{O} = cells already visited on the current path,
and \texttt{\$} = the current cell coinciding with the goal.

Generate ALL legal children of \texttt{X} in this exact order:
down, right, up, left.
A legal child is one orthogonal step inside the grid onto an
open cell (\texttt{.}) or the goal (\texttt{G}).
Never move onto a wall (\texttt{@}) or a visited cell (\texttt{O}),
outside the grid, or diagonally.

For EACH legal child, output a line containing exactly
\texttt{Push} followed by the ENTIRE child grid.
To construct each child grid:
\begin{enumerate}
  \item Start with a copy of the input grid.
  \item Mark the current cell (\texttt{X}) as visited (\texttt{O}).
  \item Mark this child's destination as the new current cell
        (\texttt{X}). If the destination is the goal (\texttt{G}),
        mark it with \texttt{\$} instead.
  \item Leave every other cell unchanged, including previously
        visited cells (\texttt{O}) and the goal (\texttt{G})
        when it is not the destination.
\end{enumerate}

If there are no legal children, return an empty response:
generate no text and end your turn.
Return only the \texttt{Push}/grid blocks, or nothing:
no explanation, coordinates, numbering, Markdown fences,
or blank lines between blocks.

An example state and its correct response follow.
Apply the same rules to the current state provided afterward.

\medskip
\textbf{User message}

Example state:
\begin{quote}
\ttfamily
@@@@@@@@@@@\\
@.OO......@\\
@..O......@\\
@.@XG.....@\\
@.........@\\
@..@@.....@\\
@.........@\\
@....@@...@\\
@.........@\\
@.........@\\
@@@@@@@@@@@
\end{quote}

\textbf{Assistant message}
\begin{quote}
\ttfamily
Push\\
@@@@@@@@@@@\\
@.OO......@\\
@..O......@\\
@.@OG.....@\\
@..X......@\\
@..@@.....@\\
@.........@\\
@....@@...@\\
@.........@\\
@.........@\\
@@@@@@@@@@@\\
Push\\
@@@@@@@@@@@\\
@.OO......@\\
@..O......@\\
@.@O\$.....@\\
@.........@\\
@..@@.....@\\
@.........@\\
@....@@...@\\
@.........@\\
@.........@\\
@@@@@@@@@@@
\end{quote}

\medskip
\textbf{User message}

Current state --- generate children for this state:
\begin{quote}
\ttfamily
@@@@@@@@@@@\\
@..G......@\\
@.@@@.@@@@@\\
@.@.@.....@\\
@@@.@@@@@.@\\
@...@.....@\\
@.@.@@@.@.@\\
@.@.@..X@.@\\
@.@@@.@.@.@\\
@.....@.@.@\\
@@@@@@@@@@@
\end{quote}

\end{tcolorbox}

\section{Extended Training Scaling Results} \label{sec: complete-training-instances}

In this section, we provide the complete results of the experiment designed to investigate how learning scales with the number of complete training instances used during training. The results for BlocksWorld and Extended BlocksWorld are presented in Table~\ref{tab:blocks-success-rates}, while the results for the Pancake Puzzle are presented in Table~\ref{tab:pancake-success-rates}. The overall findings are consistent with those reported in the main text; however, these detailed results provide a more fine-grained view of how performance changes as the number of training instances is reduced, particularly for larger values of $n$.

For TOH, we report the values of $n$ for which each hold-out setting fails, together with the corresponding failure step whenever an execution error occurs. The results are shown in Table~\ref{tab:toh_holdout_success}. In one failure case at $n=15$ with a $15\%$ hold-out setting, the entire plan was executed without any explicit execution error; however, the final state was incorrect, with only 12 disks moved to the third peg while the three largest disks remained on the first peg. For this case, the failure step is reported as ``--'' since no execution error occurred during the planning process.

\begin{table}[hbtp]
\centering
\scriptsize
\setlength{\tabcolsep}{2.5pt}
\caption{Success rates (\%) for BlocksWorld and Extended BlocksWorld across all complexity levels ($n$ blocks) and numbers of complete training instances.}
\label{tab:blocks-success-rates}

\begin{subtable}[t]{0.49\textwidth}
\centering
\caption{BlocksWorld}
\label{tab:blocksworld-success-rates}
\begin{tabular}{crrrrrrrrr}
\toprule
$n$ & \multicolumn{9}{c}{Training Instances} \\
\cmidrule(lr){2-10}
& 500 & 400 & 300 & 200 & 100 & 50 & 30 & 10 & 5 \\
\midrule
6  & 100 & 100 & 100 & 100 & 100 & 100 & 100 & 100 & 94 \\
7  & 100 & 100 & 100 & 100 & 100 & 100 & 100 & 100 & 98 \\
8  & 100 & 100 & 100 & 100 & 100 & 100 & 100 & 100 & 92 \\
9  & 100 & 100 & 100 & 100 & 100 & 100 & 100 & 100 & 84 \\
10 & 100 & 100 & 100 & 100 & 100 & 100 & 100 & 100 & 80 \\
11 & 100 & 100 & 100 & 100 & 100 & 100 & 100 & 100 & 82 \\
12 & 100 & 100 & 100 & 100 & 100 & 100 & 100 & 100 & 80 \\
13 & 100 & 100 & 100 & 100 & 100 & 100 & 100 & 100 & 84 \\
14 & 100 & 100 & 100 & 100 & 100 & 100 & 100 & 100 & 92 \\
15 & 100 & 100 & 100 & 100 & 100 & 100 & 100 & 100 & 100 \\
16 & 100 & 100 & 100 & 100 & 100 & 100 & 100 & 100 & 92 \\
17 & 100 & 100 & 100 & 100 & 100 & 100 & 100 & 100 & 90 \\
18 & 100 & 100 & 100 & 100 & 100 & 100 & 100 & 100 & 98 \\
19 & 100 & 100 & 100 & 100 & 100 & 100 & 100 & 100 & 92 \\
20 & 100 & 100 & 100 & 100 & 100 & 100 & 100 & 100 & 76 \\
21 & 100 & 100 & 100 & 100 & 100 & 100 & 100 & 96 & 82 \\
22 & 100 & 100 & 100 & 100 & 100 & 100 & 100 & 92 & 82 \\
23 & 100 & 100 & 100 & 100 & 100 & 100 & 100 & 96 & 74 \\
24 & 100 & 100 & 100 & 100 & 100 & 100 & 100 & 98 & 72 \\
25 & 100 & 100 & 100 & 100 & 100 & 100 & 100 & 98 & 78 \\
26 & 100 & 100 & 100 & 100 & 100 & 100 & 100 & 94 & 58 \\
27 & 100 & 100 & 100 & 100 & 100 & 100 & 100 & 96 & 64 \\
28 & 100 & 100 & 100 & 100 & 100 & 100 & 100 & 90 & 64 \\
29 & 100 & 100 & 100 & 100 & 100 & 100 & 100 & 94 & 52 \\
30 & 100 & 100 & 100 & 100 & 100 & 100 & 100 & 92 & 54 \\
31 & 100 & 100 & 100 & 100 & 100 & 100 & 100 & 92 & 36 \\
32 & 100 & 100 & 100 & 100 & 100 & 100 & 100 & 92 & 34 \\
33 & 100 & 100 & 100 & 100 & 100 & 100 & 100 & 92 & 40 \\
34 & 100 & 100 & 100 & 100 & 100 & 100 & 100 & 88 & 30 \\
35 & 100 & 100 & 100 & 100 & 100 & 100 & 100 & 90 & 30 \\
36 & 100 & 100 & 100 & 100 & 100 & 100 & 100 & 88 & 30 \\
37 & 100 & 100 & 100 & 100 & 100 & 100 & 100 & 94 & 20 \\
38 & 100 & 100 & 100 & 100 & 100 & 100 & 100 & 86 & 10 \\
39 & 100 & 100 & 100 & 100 & 100 & 100 & 100 & 0 & 6 \\
40 & 100 & 100 & 100 & 100 & 100 & 98 & 92 & 2 & 0 \\
\bottomrule
\end{tabular}
\end{subtable}
\hfill
\begin{subtable}[t]{0.49\textwidth}
\centering
\caption{Extended BlocksWorld}
\label{tab:extended-blocksworld-success-rates}
\begin{tabular}{crrrrrrrrr}
\toprule
$n$ & \multicolumn{9}{c}{Training Instances} \\
\cmidrule(lr){2-10}
& 500 & 400 & 300 & 200 & 100 & 50 & 30 & 10 & 5 \\
\midrule
6  & 100 & 100 & 100 & 100 & 100 & 100 & 100 & 96 & 94 \\
7  & 100 & 100 & 100 & 100 & 100 & 100 & 100 & 98 & 96 \\
8  & 100 & 100 & 100 & 100 & 100 & 100 & 100 & 86 & 88 \\
9  & 100 & 100 & 100 & 100 & 100 & 100 & 100 & 98 & 94 \\
10 & 100 & 100 & 100 & 100 & 100 & 100 & 100 & 98 & 88 \\
11 & 100 & 100 & 100 & 100 & 100 & 100 & 100 & 94 & 86 \\
12 & 100 & 100 & 100 & 100 & 100 & 100 & 100 & 88 & 82 \\
13 & 100 & 100 & 100 & 100 & 100 & 100 & 100 & 90 & 86 \\
14 & 100 & 100 & 100 & 100 & 100 & 100 & 100 & 92 & 82 \\
15 & 100 & 100 & 100 & 100 & 100 & 100 & 100 & 82 & 78 \\
16 & 100 & 100 & 100 & 100 & 100 & 100 & 100 & 78 & 80 \\
17 & 100 & 100 & 100 & 100 & 100 & 100 & 98 & 80 & 68 \\
18 & 100 & 100 & 98 & 100 & 98 & 100 & 96 & 82 & 60 \\
19 & 100 & 100 & 100 & 100 & 100 & 100 & 100 & 80 & 62 \\
20 & 100 & 100 & 100 & 100 & 100 & 100 & 100 & 76 & 72 \\
21 & 100 & 100 & 100 & 100 & 100 & 100 & 100 & 66 & 66 \\
22 & 100 & 100 & 100 & 100 & 100 & 100 & 96 & 82 & 66 \\
23 & 100 & 100 & 100 & 100 & 100 & 100 & 98 & 78 & 64 \\
24 & 100 & 100 & 100 & 98 & 98 & 100 & 98 & 74 & 70 \\
25 & 100 & 100 & 100 & 100 & 100 & 100 & 100 & 66 & 52 \\
26 & 100 & 100 & 98 & 100 & 100 & 100 & 100 & 74 & 46 \\
27 & 100 & 100 & 100 & 100 & 100 & 100 & 98 & 88 & 48 \\
28 & 100 & 98 & 100 & 100 & 100 & 96 & 98 & 60 & 36 \\
29 & 100 & 100 & 100 & 100 & 100 & 100 & 98 & 60 & 36 \\
30 & 100 & 100 & 100 & 100 & 100 & 100 & 98 & 60 & 38 \\
31 & 100 & 100 & 100 & 100 & 100 & 100 & 94 & 62 & 26 \\
32 & 100 & 100 & 100 & 100 & 100 & 96 & 94 & 66 & 22 \\
33 & 100 & 100 & 100 & 100 & 100 & 100 & 98 & 64 & 30 \\
34 & 100 & 100 & 100 & 100 & 100 & 100 & 100 & 68 & 26 \\
35 & 98 & 100 & 100 & 100 & 98 & 100 & 100 & 66 & 16 \\
36 & 98 & 100 & 100 & 100 & 98 & 96 & 100 & 54 & 16 \\
37 & 100 & 98 & 100 & 100 & 100 & 98 & 100 & 48 & 12 \\
38 & 100 & 100 & 98 & 98 & 98 & 98 & 96 & 58 & 2 \\
39 & 100 & 96 & 100 & 100 & 98 & 98 & 92 & 52 & 6 \\
40 & 100 & 96 & 98 & 100 & 98 & 94 & 94 & 40 & 2 \\
\bottomrule
\end{tabular}
\end{subtable}

\end{table}

\begin{table*}[hbtp]
\centering
\normalsize
\caption{Tower of Hanoi success results across holdout percentages. A value of 1 indicates success. A value of 0 indicates failure, with the failure step shown in parentheses.}
\label{tab:toh_holdout_success}

\resizebox{\textwidth}{!}{%
\begin{tabular}{c|cccccccccc}
\hline
& \multicolumn{10}{c}{Holdout percentage} \\
\cline{2-11}
$n$ disks 
& 5\% & 10\% & 15\% & 20\% & 25\% 
& 30\% & 35\% & 40\% & 45\% & 50\% \\
\hline
1  & 1 & 1 & 1 & 1 & 1 & 1 & 1 & 1 & 1 & 1 \\
2  & 1 & 1 & 1 & 1 & 1 & 1 & 1 & 1 & 1 & 1 \\
3  & 1 & 1 & 1 & 1 & 1 & 1 & 1 & 1 & 1 & 1 \\
4  & 1 & 1 & 1 & 1 & 1 & 1 & 1 & 1 & 0 (8) & 0 (8) \\
5  & 1 & 1 & 1 & 0 (7) & 1 & 1 & 1 & 1 & 1 & 0 (7) \\
6  & 1 & 1 & 1 & 1 & 1 & 1 & 1 & 1 & 1 & 1 \\
7  & 1 & 1 & 1 & 1 & 1 & 1 & 1 & 1 & 1 & 0 (25) \\
8  & 1 & 1 & 1 & 1 & 1 & 1 & 1 & 1 & 0 (194) & 0 (26) \\
9  & 1 & 1 & 1 & 1 & 1 & 1 & 1 & 1 & 0 (509) & 0 (27) \\
10 & 1 & 1 & 1 & 1 & 1 & 1 & 1 & 1 & 0 (197) & 0 (1152) \\
11 & 1 & 0 (387) & 1 & 1 & 1 & 0 (1537) & 1 & 1 & 0 (386) & 1 \\
12 & 1 & 0 (1827) & 1 & 1 & 0 (6146) & 0 (1538) & 1 & 0 (1828) & 0 (509) & 0 (1827) \\
13 & 1 & 1 & 1 & 1 & 1 & 0 (1539) & 1 & 0 (1830) & 0 (6144) & 0 (1537) \\
14 & 1 & 1 & 0 (1830) & 1 & 0 (24576) & 0 (1158) & 0 (12288) & 0 (1830) & 0 (197) & 0 (1537) \\
15 & 1 & 0 (37855) & 0 (--) & 0 (6145) & 0 (29984) & 0 (1158) & 0 (12288) & 0 (1830) & 0 (509) & 0 (1827) \\
\hline
Success rate 
& 100.0\% 
& 80.0\% 
& 86.7\% 
& 86.7\% 
& 80.0\% 
& 66.7\% 
& 86.7\% 
& 73.3\% 
& 40.0\% 
& 33.3\% \\
\hline
\end{tabular}%
}
\end{table*}

\begin{table}[t]
\centering
\scriptsize
\setlength{\tabcolsep}{3pt}
\caption{Success rates (\%) for Pancake across all complexity levels
($n$ pancakes) and numbers of complete training instances.}
\label{tab:pancake-success-rates}
\begin{tabular}{crrrrrrrrr}
\toprule
$n$ & \multicolumn{9}{c}{Training Instances} \\
\cmidrule(lr){2-10}
& 500 & 400 & 300 & 200 & 100 & 50 & 30 & 10 & 5 \\
\midrule
6  & 100 & 100 & 100 & 100 & 100 & 100 & 98 & 88 & 78 \\
7  & 100 & 100 & 100 & 100 & 100 & 100 & 94 & 88 & 66 \\
8  & 100 & 100 & 100 & 100 & 100 & 100 & 96 & 94 & 48 \\
9  & 100 & 100 & 100 & 100 & 100 & 100 & 98 & 88 & 8 \\
10 & 100 & 100 & 100 & 100 & 100 & 100 & 96 & 88 & 66 \\
11 & 100 & 100 & 100 & 100 & 100 & 100 & 100 & 84 & 40 \\
12 & 100 & 100 & 100 & 100 & 100 & 100 & 100 & 76 & 22 \\
13 & 100 & 100 & 100 & 100 & 100 & 100 & 100 & 68 & 48 \\
14 & 100 & 100 & 100 & 100 & 100 & 100 & 98 & 76 & 60 \\
15 & 100 & 100 & 100 & 100 & 100 & 100 & 100 & 70 & 74 \\
16 & 100 & 100 & 100 & 100 & 100 & 100 & 100 & 72 & 84 \\
17 & 100 & 100 & 100 & 100 & 100 & 100 & 100 & 82 & 44 \\
18 & 100 & 100 & 100 & 100 & 100 & 100 & 100 & 76 & 84 \\
19 & 100 & 100 & 100 & 100 & 100 & 100 & 100 & 76 & 86 \\
20 & 100 & 100 & 100 & 100 & 100 & 100 & 100 & 52 & 54 \\
21 & 100 & 100 & 100 & 100 & 100 & 100 & 98 & 58 & 92 \\
22 & 100 & 100 & 100 & 100 & 100 & 100 & 100 & 86 & 90 \\
23 & 100 & 100 & 100 & 100 & 100 & 100 & 100 & 76 & 46 \\
24 & 100 & 100 & 100 & 100 & 100 & 100 & 100 & 76 & 44 \\
25 & 100 & 100 & 100 & 100 & 100 & 100 & 100 & 56 & 64 \\
26 & 100 & 100 & 100 & 100 & 100 & 100 & 100 & 74 & 56 \\
27 & 100 & 100 & 100 & 100 & 100 & 100 & 100 & 52 & 76 \\
28 & 100 & 100 & 100 & 100 & 100 & 100 & 100 & 68 & 54 \\
29 & 100 & 100 & 100 & 100 & 100 & 100 & 100 & 58 & 44 \\
30 & 100 & 100 & 100 & 100 & 100 & 100 & 98 & 70 & 48 \\
31 & 100 & 100 & 100 & 100 & 100 & 100 & 100 & 70 & 32 \\
32 & 100 & 100 & 100 & 100 & 100 & 100 & 100 & 66 & 46 \\
33 & 100 & 100 & 100 & 100 & 100 & 100 & 100 & 20 & 54 \\
34 & 100 & 100 & 100 & 100 & 98 & 100 & 98 & 62 & 20 \\
35 & 100 & 100 & 100 & 100 & 100 & 100 & 100 & 58 & 20 \\
36 & 100 & 100 & 100 & 100 & 100 & 100 & 100 & 62 & 6 \\
37 & 100 & 100 & 100 & 100 & 100 & 100 & 100 & 58 & 10 \\
38 & 100 & 100 & 100 & 100 & 100 & 94 & 100 & 36 & 0 \\
39 & 100 & 100 & 100 & 100 & 96 & 100 & 96 & 30 & 2 \\
40 & 100 & 100 & 100 & 100 & 100 & 98 & 82 & 18 & 0 \\
\bottomrule
\end{tabular}
\end{table}

\section{Context Window Growth during Planning} \label{sec: context window growth}

\begin{figure}[h]
    \centering
    \includegraphics[width=\textwidth]{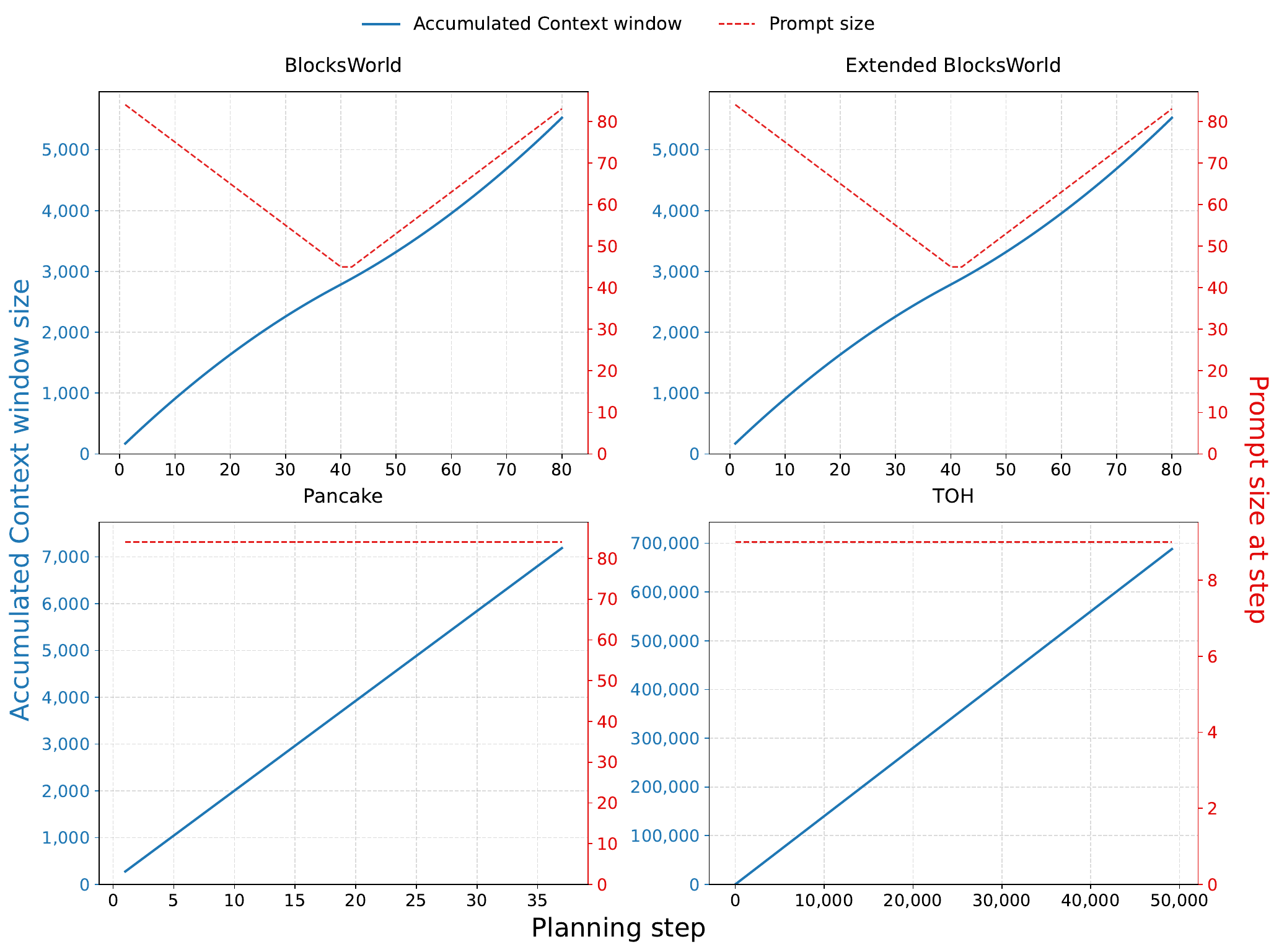}
    \caption{
    Prompt size and cumulative context-window size over planning steps for BlocksWorld,
    Extended BlocksWorld, Pancake, and Tower of Hanoi (TOH). The x-axis shows the
    planning step. The context-window curve measures the initial prompt plus all
    model-generated text up to that step, while the prompt-size curve measures the
    prompt at the current step.}
    \label{fig:context-window-growth}
\end{figure}

In this section, we examine how the context window evolves throughout the planning process across all environments. In this experiment, the context window is defined as the initial prompt together with all targets generated by the language model and written to the tape throughout planning. Since the highest complexity level in each environment exhibits the longest planning trajectory and subsumes all planning patterns observed at lower complexity levels, we report results only for the largest value of $n$ in each domain. Figure~\ref{fig:context-window-growth} presents the results for all four environments, showing both the context window size and the prompt size at each planning step. As shown in the figure, the size of the prompt passed to the language model remains nearly constant throughout planning for TOH and the Pancake Puzzle, while the total context window grows linearly with the solution length. In both versions of BlocksWorld, the prompt size initially decreases as the start configuration is progressively unstacked. (Single blocks on the table are not explicitly represented, but are implicitly found on the table.) The minimum prompt size is reached when all blocks are placed on the table. Thereafter, the prompt size increases as the blocks are stacked to match the goal configuration.

These results highlight the effectiveness of the pointer mechanism for planning. First, it helps maintain prediction accuracy throughout the planning process. Once the model has learned to execute a particular instruction, its accuracy is largely independent of the total solution length because it only needs to attend to the current instruction and the relevant local information. In contrast, if the entire accumulated context were provided as input at every step, the model would need to retrieve the relevant information from an increasingly large context window, making the task progressively more difficult and potentially reducing accuracy. Second, the pointer mechanism substantially reduces computational cost. In our approach, the amount of information processed by the language model at each planning step remains approximately constant. In contrast, when the entire accumulated context is repeatedly passed to the model, the computational cost increases with the context length, since all tokens in the context participate in the attention computation. As a result, the overall computational cost grows with the length of the planning trajectory, whereas in our approach the cost per planning step remains nearly constant.

%% file: sections/related_work.tex
\section{Related Work} \label{sec: realted work}

\paragraph{Overview of Research Directions in LLM Planning}
We broadly categorize prior work at the intersection of LLMs and planning into several groups. First, survey papers provide overviews of existing approaches and outline directions for future research \citep{huang2025chasing,chiari2025planning,tantakoun2025llms,pallagani2024prospects}. Second, empirical and benchmark studies evaluate and compare the planning abilities of state-of-the-art LLMs across different domains \citep{dziri2023faith,shojaee2025illusion,kokel2025acpbench,kokel2026acpbench,pallagani2023understanding,valmeekam2023planning,aghzal2024pathplanners,bohnet2024exploring,liang2026hardcorelogic,bohnet2026analysis}. Third, a line of work focuses on integrating LLMs with planning algorithms \citep{meng2024llm,hazra2024saycanpay,hao2023reasoning,yao2023tree,shah2023navigation,katz2024thought,schultz2025mastering,li2025ground,correa2025classical} or using search-based reasoning (e.g., tree search) within LLM prompting \citep{sel2024algorithm,gandhi2024stream}. Finally, theoretical works study the computational properties of LLMs, including their general computation capabilities \citep{lewandowski2026universal,malach2024autoregressive,schuurmans2023memory,jiang2026softmax,li2024chain,hou2025universal}, as well as their ability to learn algorithms with guarantees, often through the lens of length generalization \citep{giannou2023looped,zhou2024algorithms,hua2026imitation,malach2026infinity,chen2026canllms,fan2025looped}.

\paragraph{LLM-Augmented Planning Methods} 


\citet{meng2024llm} propose a planning algorithm, \emph{LLM-A*}, in which the classical A* algorithm is augmented with an LLM. The model is prompted to generate a sequence of intermediate target states at the beginning of planning, such that the sequence starts from the initial state and ends at the goal state. These intermediate targets guide the search, and an additional cost term is introduced based on the distance to the current target. \citet{hao2023reasoning} propose \emph{Reasoning via Planning} (RAP), which integrates LLMs with an offline planning algorithm such as Monte Carlo Tree Search (MCTS). In this framework, the LLM plays two roles: (1) as a world model that generates new states in the search tree, and (2) as a policy that selects actions during simulations. To reduce the number of queries required by multi-step reasoning methods, \citet{sel2024algorithm} propose \emph{Algorithm of Thoughts} (AoT), an in-context learning approach in which the model is provided with examples of the search process itself, rather than only problem--solution pairs or step-by-step solutions.

\paragraph{Turing-Completeness of LLMs} 


\citet{schuurmans2023memory} show that the fundamental limits of LLM reasoning are not necessarily rooted in their weights, but rather in the constraints imposed by the context window and available memory. They prove that transformer-based LLMs are Turing-complete, i.e., capable of universal computation, when equipped with an external read--write memory. \citet{lewandowski2026universal} further show that the context window itself can serve as the available memory for the LLM. Specifically, they demonstrate that an LLM can emulate a tag system, which, by moving both forward and backward over the context, can simulate a Turing machine. \citet{malach2024autoregressive} argue that the power of modern LMs largely stems from their autoregressive next-token prediction training, particularly when applied to Chain-of-Thought (CoT) reasoning. They show that even linear next-token predictors can approximate any function that is efficiently computable by a Turing machine. \citet{jiang2026softmax} identify a key limitation in prior proofs of Turing-completeness for CoT-based transformers \citep{li2025constant,meng2024llm}, namely the reliance on hardmax attention, which is not a realistic assumption and does not provide guarantees on trainability. To address this limitation, they prove that transformers with standard softmax attention are also Turing-complete and provide guarantees on length generalization.

\paragraph{Length-generalization of LLMs.} 


\citet{zhou2024algorithms} study which algorithms are easier for transformers to learn. They characterize an algorithm as easier to learn (e.g., counting or mode computation) if the model can length-generalize, i.e., generalize to longer instances of the same problem. They further show that tasks expressible in the RASP-L language, a learnable variant of the original RASP language \citep{weiss2021thinking}, tend to exhibit better length generalization, although such generalization is not guaranteed and remains limited in practice. \citet{malach2026infinity} show that, although state space models (SSMs) \citep{gu2022efficiently} are more computationally efficient for long contexts compared to transformers (whose complexity scales quadratically with sequence length), they still fail to solve problems when the required context exceeds the model's capacity. They further demonstrate that incorporating external tool use significantly improves the length generalization of both SSMs and transformers. \citet{hua2026imitation} propose Turing Machine Imitation Learning (TAIL), where the goal is to imitate the serialized linear state transitions of a Turing machine through COT reasoning. TAIL consists of three core components: (1) \emph{Linear Transition}, which enforces sequential computation; (2) \emph{Atomic States}, which ensures that each step is sufficiently simple to avoid shortcut learning; and (3) \emph{Memory Fetcher}, which provides the necessary operands for each transition. Although TAIL improves length generalization across a variety of computable problems, there remains a performance gap compared to strong closed-source reasoning models (e.g., O4-mini) when evaluated on models such as Qwen2.5 \citep{hui2024qwen2}.

\paragraph{Limitations of LLMs in Planning} 


\citet{valmeekam2023planning} were among the first to systematically investigate claims about the reasoning abilities of LLMs by evaluating their performance on planning tasks. Their findings indicate that LLMs at the time were rarely successful at autonomously generating valid solution plans. Notably, they show that performance degrades significantly when object names are obfuscated, suggesting that LLM-generated plans rely more on learned semantic associations than on robust reasoning. Building on this line of work, \citet{shojaee2025illusion} revisit similar experiments using newer reasoning-oriented models. They find that while reasoning models can outperform non-reasoning models on instances of moderate complexity, both types of models fail at similar levels as problem complexity increases. \citet{katz2024thought} highlight the computational cost and inefficiency of LLM-based planning approaches such as RAP \citep{hao2023reasoning}. They show that a common limitation of these methods is the need to query the LLM at every step of the planning process, which becomes prohibitively slow and costly in large state spaces. To address this, they propose using LLMs to generate environment simulators and associated functions prior to planning, which can then be leveraged by classical planning algorithms.

\paragraph{Reasoning Investigation in LLMs} 


\citet{dziri2023faith} identify two key challenges that motivate our approach. First, as problem complexity grows, errors introduced in early stages of the computation tend to accumulate, leading to degraded performance on longer and more complex instances. Second, LLMs often exhibit shortcut learning \citep{geirhos2020shortcut}, relying on pattern matching over linearized solution traces rather than reasoning and learning the underlying compositional structure of the task. \citet{chen2026canllms} study the reasoning abilities of LLMs in a high-stakes domain, namely legal judgment. They analyze the representations and interactions between input tokens to assess their influence on generated outputs. One of their key findings is that LLMs often base their decisions on semantically irrelevant phrases, raising concerns about their reasoning capabilities.